\PassOptionsToPackage{table,xcdraw}{xcolor}
\documentclass{article}

\usepackage{iclr2026_conference,times}
\usepackage{subcaption}

\newcommand{\ind}[3]{_{#1 = #2}^{#3}}

\newcommand{\trsp}{^{\top}}
\newcommand{\diag}{\mathrm{diag}}

\def\e{\bm{e}}
\def\x{\bm{x}}
\def\y{\bm{y}}

\def\M{\bm{M}}

\def\S{\bm{S}}

\def\W{\bm{W}}

\def\1{\bm{1}}
\def\0{\bm{0}}

\newcommand{\mcA}{\mathcal{A}}
\newcommand{\mcD}{\mathcal{D}}
\newcommand{\mcK}{\mathcal{K}}
\newcommand{\mcL}{\mathcal{L}}
\newcommand{\mcM}{\mathcal{M}}

\newcommand{\mcQ}{\mathcal{Q}}

\newcommand{\mcS}{\mathcal{S}}
\newcommand{\mcF}{\mathcal{F}}

\newcommand{\mcV}{\mathcal{V}}

\newcommand{\mcG}{\mathcal{G}}

\newcommand{\mcY}{\mathcal{Y}}
\newcommand{\mcH}{\mathcal{H}}

\newcommand{\mbD}{\mathbb{D}}

\newcommand{\mbR}{\mathbb{R}}

\newcommand{\mbN}{\mathbb{N}}

\usepackage{amsmath}
\usepackage{amsfonts}
\usepackage{amsthm}
\usepackage{amssymb}
\usepackage{fontawesome5}
\usepackage{pgffor}
\usepackage{multirow}

\usepackage{microtype}
\usepackage{graphicx}
\usepackage{booktabs} 

\usepackage{graphicx}
\usepackage{caption}
\usepackage{subcaption}
\usepackage{bm}
\usepackage{amsmath}
\usepackage{amssymb}
\usepackage{amsthm}
\usepackage{wrapfig}
\usepackage{comment}
\usepackage{enumitem}
\usepackage[misc]{ifsym}
\usepackage{diagbox}

\usepackage{mathtools}

\usepackage{mathrsfs}

\usepackage[utf8]{inputenc} 
\usepackage[T1]{fontenc}    
\usepackage{url}            
\usepackage{booktabs}       
\usepackage{amsfonts}       
\usepackage{nicefrac}       
\usepackage{microtype}      
\usepackage{tabularx}
\usepackage[misc]{ifsym}

\usepackage{pifont}

\usepackage[pagebackref=true,breaklinks=true,colorlinks=true,bookmarks=false]{hyperref}
\definecolor{DirtyOrange}{HTML}{D87C46}
\hypersetup{linkcolor=DirtyOrange}
\hypersetup{urlcolor=DirtyOrange}
\definecolor{Indigo}{HTML}{4B0082}
\hypersetup{citecolor=Indigo}

\usepackage{hyperref}

\newtheorem{thm}{Theorem}
\newtheorem{lem}[thm]{Lemma}
\newtheorem{prop}[thm]{Proposition}

\newtheorem{defn}[thm]{Definition}

\usepackage[algo2e,algoruled,boxed,lined]{algorithm2e}
\newcommand{\ie}{\emph{i.e.}}
\newcommand{\eg}{\emph{e.g.}}
\usepackage{ragged2e}

\newcommand{\revise}[1]{#1}

\definecolor{myGreen}{HTML}{33CC33}

\definecolor{myPurple}{HTML}{9966FF}

\definecolor{myOrange}{HTML}{FF9900}

\usepackage{minitoc}

\usepackage[table,xcdraw]{xcolor} 
\usepackage{algorithm}
\usepackage{titletoc} 
\usepackage{academicons}
\newif\ifsubmission
\submissionfalse

\ifsubmission
\newcommand{\mcnote}[1]{}
\else
\usepackage[textsize=small]{todonotes}
\newcommand{\mcnote}[1]{\todo[color=purple!40,inline]{MC: #1}}
\fi

\title{Nonparametric Teaching of Attention \\ Learners}

\author{\begin{tabular}{c}
    Chen Zhang\textsuperscript{1}\thanks{Equal contribution} \quad 
    Jianghui Wang\textsuperscript{2}\footnote[1]{} \quad 
    Bingyang Cheng\textsuperscript{1} \quad Zhongtao Chen\textsuperscript{1} \quad Wendong Xu\textsuperscript{1}
    \\[2pt]  
    Cong Wang\textsuperscript{3} \quad 
    Marco Canini\textsuperscript{2} \quad 
    Francesco Orabona\textsuperscript{2} \quad
    Yik-Chung Wu\textsuperscript{1} \quad Ngai Wong\textsuperscript{1}
  \end{tabular}\\
\\
\textsuperscript{1} The University of Hong Kong \\
\textsuperscript{2} King Abdullah University of Science and Technology \\
\textsuperscript{3} Independent Researcher \\
\texttt{czhang6@connect.hku.hk}\quad\texttt{jianghui.wang@kaust.edu.sa}
\\[8pt]
  \centerline{%
    \raisebox{-0.2\height}{\includegraphics[height=1.3em]{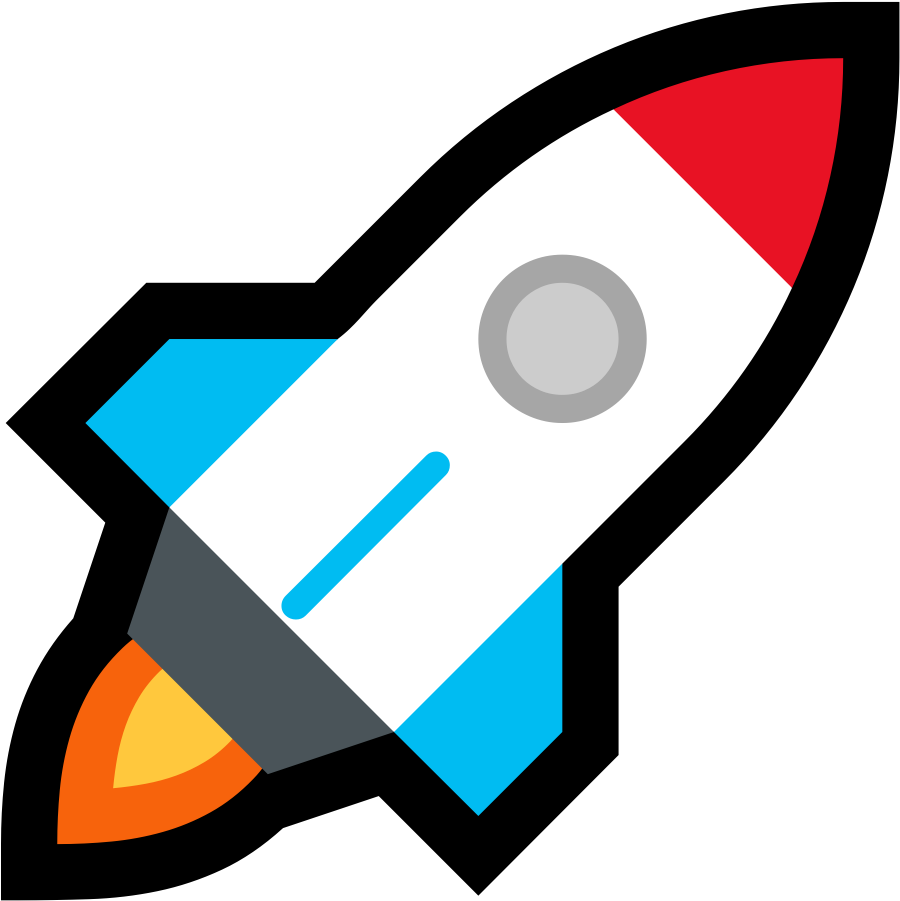}}\ \href{https://chen2hang.github.io/_publications/nonparametric_teaching_of_attention_learners/attent.html}{\texttt{\large Project page}}   
  }
}

\iclrfinalcopy

\begin{document}

\maketitle

\begin{abstract}
Attention learners, neural networks built on the attention mechanism, \eg, transformers, excel at learning the implicit relationships that relate sequences to their corresponding properties, \eg, mapping a given sequence of tokens to the probability of the next token. However, the learning process tends to be costly. To address this, we present a novel paradigm named \textbf{Atte}ntion \textbf{N}eural \textbf{T}eaching (AtteNT) that reinterprets the learning process through a nonparametric teaching perspective. Specifically, the latter provides a theoretical framework for teaching mappings that are implicitly defined (\ie, nonparametric) via example selection. Such an implicit mapping is embodied through a dense set of sequence-property pairs, with the AtteNT teacher selecting a subset to accelerate convergence in attention learner training. By analytically investigating the role of attention on parameter-based gradient descent during training, and recasting the evolution of attention learners, shaped by parameter updates, through functional gradient descent in nonparametric teaching, we show \textit{for the first time} that teaching attention learners is consistent with teaching importance-adaptive nonparametric learners. 
These new findings readily commit AtteNT to enhancing learning efficiency of attention learners. Specifically, we observe training time reductions of 13.01\% for LLMs and 20.58\% for ViTs, spanning both fine-tuning and training-from-scratch regimes. Crucially, these gains are achieved without compromising accuracy; in fact, performance is consistently preserved and often enhanced across a diverse set of downstream tasks.

\end{abstract}

\section{Introduction}

The attention mechanism, inspired by human attention concepts \citep{ahmad1991visit,soydaner2022attention}, is designed to assess the relative importance of each element in a sequence \citep{bahdanau2015neural,vaswani2017attention}. By leveraging attention, neural networks can effectively learn the implicit relationships that map sequences to their corresponding properties, \eg, mapping a sequence of tokens to the probability of the next token. These Attention Neural Networks (ANNs), \eg, transformers \citep{vaswani2017attention,kong2019weakly}, have achieved significant success in a wide range of downstream tasks across various fields, including natural language processing \citep{vaswani2017attention,devlin2019bert,consens2025transformers}, computer vision \citep{dosovitskiyimage,azad2024beyond,chen2024dnnam}, and multimodal systems \citep{nagrani2021attention,yang2024mran}.

However, the process of learning the implicit mappings (\ie, training) can be quite costly for ANNs, especially when handling large-scale tasks~\citep{liu2018lcqmc,beltagy2019scibert,gu2021domain,yang2023vid2seq}. For instance, pretraining language models often requires training on corpora with millions of sentences \citep{commoncrawl2007,li2024datacomp}. In the case of video understanding, the scale can become overwhelmingly large \citep{bain2021frozen,sharma2018conceptual,shu2025video}. As a result, reducing training costs and enhancing learning efficiency has become an urgent priority.

Recent research on nonparametric teaching~\citep{zhang2023nonparametric, zhang2023mint,zhang2024nonparametric,zhang2025nonparametric} presents a promising solution to the issue outlined above. Specifically, nonparametric teaching provides a theoretical framework for selecting examples efficiently when the target mapping (\ie, either a function or a model) being taught is nonparametric, \ie, implicitly defined. It builds on the concept of machine teaching~\citep{zhu2015machine, zhu2018overview}, which involves designing a training set (dubbed the teaching set) to help the learner quickly converge to the target functions, while relaxing the assumption that the target functions are parametric~\citep{liu2017iterative,liu2018towards}, thus enabling the teaching of nonparametric (non-closed-form) functions with a focus on function space. Unfortunately, these studies are limited to multilayer perceptron-based learners and do not account for the attention mechanism, making their direct application difficult when the learners are ANNs.  Additionally, ANNs are typically updated through gradient descent in parameter space, which contrasts with the functional gradient descent used in nonparametric teaching within function space~\citep{zhang2023nonparametric, zhang2023mint, zhang2024nonparametric, zhang2025nonparametric}. Hence, it is not immediate to apply nonparametric teaching theory to attention learners.

To this end, we systematically investigate the role of attention on ANN gradient-based training in both parameter and function spaces. Specifically, we analytically examine how attention adaptively assigns different importance to each element in an input sequence during parameter-based gradient descent in parameter space, and explicitly show that the parameter gradient retains the same form as the input sequence size scales. This importance-adaptive update in parameter space drives the evolution of the ANN, which can be expressed using the dynamic Attention Neural Tangent Kernel (ANTK)~\citep{yang2019wide,hron2020infinite}, and then cast into function space. We prove that this dynamic ANTK converges to the importance-adaptive canonical kernel used in functional gradient descent, suggesting that the evolution of ANN under parameter gradient descent is consistent with that under functional gradient descent. Therefore, it is natural to interpret the learning process of attention learners through the theoretical framework of nonparametric teaching: the target mapping is represented by a dense set of sequence-property pairs, where each sequence is associated with its target output, and the teacher selects a subset of these pairs to provide to the ANN, ensuring rapid convergence of this attention learner. Consequently, to improve the learning efficiency of ANNs, we propose a novel paradigm called AtteNT, where the teacher applies a counterpart of the greedy teaching algorithm from nonparametric teaching to train attention learner, specifically by selecting the sequence with the greatest discrepancy between their true property values and the ANN outputs. Lastly, we carry out comprehensive experiments to demonstrate the effectiveness of AtteNT across various scenarios, including both natural language processing and computer vision tasks. Our key contributions are as follows:
\begin{itemize}[leftmargin=*,nosep]
	\setlength\itemsep{0.56em}
	\item We propose AtteNT, a novel paradigm that interprets attention learner training through the theoretical lens of nonparametric teaching. This  facilitates the use of greedy algorithms from nonparametric teaching to effectively improve the learning efficiency of attention learners.
	\item We analytically investigate the role of attention in parameter-based gradient descent within parameter space, revealing the consistency between the evolution of ANN driven by parameter updates and that under functional gradient descent in nonparametric teaching. We further show that the dynamic ANTK, emerging from gradient descent on the parameters, converges to the importance-adaptive canonical kernel of functional gradient descent. These findings bridge nonparametric teaching theory with attention learner training, thereby broadening the application of nonparametric teaching to contexts involving attention mechanisms.
	\item We demonstrate the effectiveness of AtteNT through extensive experiments across both natural language processing (NLP) and computer vision (CV) tasks. Our approach reduces Large Language Model (LLM) fine-tuning time by 13.01\% and accelerates Vision Transformer (ViT) training from scratch by 20.58\%, thereby providing strong empirical support for our theoretical claims.
\end{itemize}

\section{Related Works}

\textbf{Attention learners}. The effectiveness of the attention mechanism in learning implicit mappings from sequences to relevant properties has spurred a surge in research on attention learners~\citep{bahdanau2015neural, vaswani2017attention, kong2019weakly}. This growing interest is particularly evident in the increasing efforts to apply attention learners across a wide variety of downstream tasks, including natural language processing~\citep{galassi2020attention, jin2024impact}, computer vision~\citep{dosovitskiyimage, hassanin2024visual, zhang2024you}, medicine~\citep{thirunavukarasu2023large, demszky2023using}, and graph-related fields~\citep{velivckovic2018graph, wu2024survey}. Various efforts have been made in designing learners for improved mapping learning in vision tasks~\citep{lin2022survey, dosovitskiyimage, arnab2021vivit}, as well as for more efficient inference~\citep{kitaevreformer, katharopoulos2020transformers, lu2021soft}. There have also been ongoing pursuits to enhance learning efficiency, such as sparse training~\citep{frankle2018lottery, you2020drawing, chen2021earlybert, chen2021chasing, li2023lazy}, improved initialization~\citep{huang2020improving, d2021convit}, and data curation~\citep{tang2023dynamic, zhong2023revisiting, lin2024not, li2024datacomp}. Differently, we frame attention learner training from a fresh perspective of nonparametric teaching~\citep{zhang2023nonparametric,zhang2023mint}, and adopt a corresponding variant of the greedy algorithm to enhance the training efficiency of ANNs.

\textbf{Nonparametric teaching}. Machine teaching~\citep{zhu2015machine, zhu2018overview} focuses on designing a teaching set that allows the learner to quickly converge to a target model function. It can be seen as the reverse of machine learning: while machine learning aims to learn a mapping from a given training set, machine teaching seeks to construct the set based on a desired mapping. Its effectiveness has been demonstrated across various domains, including crowdsourcing~\citep{singla2014near, zhou2018unlearn}, robustness~\citep{alfeld2017explicit, ma2019policy, rakhsha2020policy}, and computer vision~\citep{wang2021gradient, wang2021machine}. Nonparametric teaching~\citep{zhang2023nonparametric, zhang2023mint} extends iterative machine teaching~\citep{liu2017iterative, liu2018towards} by broadening the parameterized family of target mappings to encompass the more general nonparametric framework. This theoretical framework has proven effective in enhancing the efficiency of multilayer perceptrons for learning implicit functions from signal coordinates to corresponding values~\citep{zhang2024nonparametric, zhang2026nint}, as well as improving the training efficiency of graph convolutional networks for learning implicit mappings from graphs to their relevant properties~\citep{zhang2025nonparametric}. Nevertheless, the absence of the attention mechanism in these studies limits their direct applicability to general tasks involving attention learners~\citep{bahdanau2015neural, vaswani2017attention}. This work systematically investigates the role of attention and highlights the alignment between the evolution of ANN driven by parameter updates and that guided by functional gradient descent in nonparametric teaching. These insights, for the first time, broaden the scope of nonparametric teaching in attention learner training, positioning our AtteNT as a novel approach to improving ANN learning efficiency.

\section{Background}
\vspace{-2pt}

\textbf{Notation}.\footnote{The notation table can be found in Appendix~\ref{table:notation}.} Let $(\x_1,\dots,\x_S)$ represent a sequence of length $S$, where each $\x_s\in \mbR^d$ denotes a $d$-dimensional feature vector  associated with the $s$-th element, with $s \in \mbN_S$ ($\mbN_S \coloneqq \{1, \dots, S\}$). Each $\x_s$ is a row vector, expressed as $[x_{s,j}]_d\trsp = [x_{s,1}, \dots, x_{s,d}]$. The entire collection of feature vectors forms an $S \times d$ feature matrix, denoted $\S_{S \times d}\in\mcS\subseteq\mbR^{S\times d}$ (or simply $\S$). The $s$-th row and the $i$-th column of this matrix, corresponding to the $s$-th element and the $i$-th feature, are denoted by $\S_{(s,:)}$ and $\S_{(:,i)}$, respectively. Alternatively, these can be written as $\e_s\trsp\S$ and $\S\e_i$, where $\e_i$ is a standard basis vector with its $i$-th entry being $1$ and all other entries equal to $0$. The bold column vector $\1$ represents a vector in which all elements are $1$. The property of the sequence is represented by $\y \in \mcY$, where $\y$ is a scalar for sequence-level properties ($\mcY \subseteq \mbR$) and a vector for element-level properties ($\mcY \subseteq \mbR^n$). A set with $m$ items is denoted as $\{a_i\}_m$. If $\{a_i\}_m \subseteq \{a_i\}_n$, then $\{a_i\}_m$ represents a subset of $\{a_i\}_n$ containing $m$ items, where the indices are $i \in \mbN_n$. A diagonal matrix with diagonal entries $a_1, \dots, a_m$ is denoted as $\diag(a_1, \dots, a_m)$, and if all $m$ values are identical,  the matrix is simplified as $\diag(a; m)$.

Let $K(\S,\S'): \mcS\times\mcS\mapsto\mbR$ denote a symmetric and positive definite sequence kernel \citep{cancedda2003word,kiraly2019kernels}. This kernel can also be expressed as $K(\S,\S')=K_{\S}(\S')=K_{\S'}(\S)$, where for simplicity, $K_{\S}(\cdot)$ may be abbreviated as $K_{\S}$. The reproducing kernel Hilbert space (RKHS) $\mcH$ associated with $K(\S,\S')$ is defined as the closure of the linear span $\{f:f(\cdot)=\sum\ind{i}{1}{r}a_i K(\S_i,\cdot),a_i\in\mbR,r\in\mbN,\S_i\in\mcS\}$,  with the inner product given by $\langle f,g\rangle_\mcH=\sum_{ij}a_ib_j K(\S_i,\S_j)$, where $g=\sum_{j}b_j K_{\S_j}$~\citep{liu2016stein,zhang2023nonparametric}. Rather than assuming the idealized case of a closed-form solution $f^*$, we focus on the more realistic scenario where the realization of $f^*$ is given~\citep{zhang2023nonparametric,zhang2023mint,zhang2024nonparametric,zhang2025nonparametric}. Given the target mapping $f^*:\mcS\mapsto\mcY$, it uniquely maps each sequence $\S_\dagger$ to its corresponding output $\y_\dagger$, such that $\y_\dagger=f^*(\S_\dagger)$. According to the Riesz–Fréchet representation theorem \citep{lax2014functional, scholkopf2002learning, zhang2023nonparametric}, the evaluation functional is defined as follows:
\begin{defn}
	\label{efl}
	Let $\mcH$ denote a reproducing kernel Hilbert space\footnote{In nonparametric teaching, the extension from scalar-valued to vector-valued functions, relating to element-level properties, is a well-established generalization in \citealp{zhang2023mint}.} equipped with a positive definite sequence kernel $K_{\S}\in\mcH$, where $\S\in\mcS$. The evaluation functional $E_{\S}(\cdot):\mcH\mapsto\mbR$ is defined by the reproducing property as
	\begin{equation}
		E_{\S}(f)=\langle f, K_{\S}(\cdot)\rangle_\mcH=f(\S), \quad f\in\mcH~.
	\end{equation}
\end{defn}
Furthermore, for a functional $F:\mcH\mapsto\mbR$, the Fréchet derivative~\citep{coleman2012calculus, liu2017stein,zhang2023nonparametric} of $F$ is defined as:
\begin{defn} (Fréchet derivative in RKHS)
	\label{fd}
	The Fréchet derivative of a functional $F:\mcH\mapsto\mbR$ at a point $f\in\mcH$, represented as $\nabla_f F(f)$, is defined implicitly by $F(f+\epsilon g)=F(f)+\langle\nabla_f F(f),\epsilon g\rangle_\mcH+o(\epsilon)$ for any $g\in\mcH$ and $\epsilon\in\mbR$. This derivative itself is a function in $\mcH$.
\end{defn}
\textbf{Attention learners}, referring to neural networks that incorporate attention mechanisms, are designed to learn the implicit mapping between input sequences and their associated properties \citep{vaswani2017attention}. Specifically, the attention consists of three components: the query matrix $\mcQ(\S)\coloneqq\S\W^{Q}$, the key matrix $\mcK(\S)\coloneqq\S\W^{K}$, and the value matrix $\mcV(\S)\coloneqq\S\W^{V}$, where the query and key weight matrices $\W^{Q}$ and $\W^{K}$ are of size $d\times p$, and the value weight matrix $\W^{V}$ is of size $d\times v$. For simplicity, this paper primarily focuses on a single-layer, single-head self-attention neural network\footnote{This can be directly extended to other attention learners, including those with multi-head attention or different types of attention mechanisms \citep{dong2021attention,kajitsuka2024transformers}.} \citep{mahankali2024one,makkuva2025attention}, which can be expressed as 
\begin{eqnarray}
f_\theta(\S)=\mathrm{softmax}\left(\frac{\mcQ(\S)\mcK(\S)\trsp}{\sqrt{d}}\right)\mcV(\S),
\end{eqnarray}
where $\mathrm{softmax}(\cdot)$ is applied row-wise.

\textbf{Nonparametric teaching} is formulated as a functional minimization over a teaching set, denoted as $\mcD=\{(\x^1,y^1),\dots(\x^T,y^T)\}$, where each input $\x\in\mbR^d$ represents independent feature vectors, without considering the sequence~\citep{zhang2023nonparametric}. The collection of all possible teaching sets is represented by $\mbD$:
\begin{eqnarray}\label{eq1}
    \mcD^*=\underset{\mcD\in\mbD}{\arg\min}\ \mcM(\hat{f},f^*)+\lambda\cdot \mathrm{card}(\mcD) \qquad\qquad\text{s.t.}\quad\hat{f}\coloneqq\mcA(\mcD)~.
\end{eqnarray}
This formulation involves three key components: $\mcM$ which measures the discrepancy between $\hat{f}$ and $f^*$ (\eg, $L_2$ distance in RKHS $\mcM(\hat{f^*},f^*)=\|\hat{f^*}-f^*\|_\mcH$); $\mathrm{card}(\cdot)$, representing the cardinality (or size) of the teaching set $\mcD$, controlled by a regularization constant $\lambda>0$; and $\mcA(\mcD)$, which denotes the learning algorithm employed by the learners, typically based on empirical risk minimization:
\begin{equation}
	\label{eq:la}
	\mcA(\mcD)\coloneqq\underset {f\in\mcH}{\arg\min}\,\frac{1}{\mathrm{card}(\mcD)}\sum_{(\x^t,y^t)\in\mcD}\mcL\big(f(\x^t),y^t\big).
\end{equation}
with a convex loss $\mcL$ (w.r.t. $f$), which is optimized using functional gradient descent:\footnote{The functional gradient is obtained by applying the functional chain rule (Lemma~\ref{cr}) and the gradient of an evaluation functional (Lemma~\ref{ef}), both of which are detailed in Appendix~\ref{app:fg}.}
\begin{equation}
	\label{opta}
	f^{t+1}\gets f^t-\eta \underbrace{E_{\x}\left(\frac{\partial\mcL(f^*,f^t)}{\partial f^t}\right)\cdot K_{\x}}_{\coloneqq\mcG(\mcL,f^*;f^t,\x)\text{, Functional Gradient}},
\end{equation}
where $t=0,1,\dots,T$ is the iteration index, $\eta>0$ is the learning rate, and $E_{\x}(f)=f(\x)$ denotes the evaluation functional.

\section{AtteNT}

We begin by investigating the role of attention in parameter-based gradient descent. Then, by translating the evolution of an ANN—driven by importance-adaptive updates in parameter space—into function space, we show that the evolution of the ANN under parameter gradient descent is consistent with that under functional gradient descent. Lastly, we present the greedy AtteNT algorithm, which effectively selects sequences with steeper gradients to enhance the learning efficiency of the ANN.

\subsection{Importance-adaptive Update in the Parameter Space}  \label{emlp}

\begin{wrapfigure}{r}{0.5\textwidth}
\vskip -1mm
\centering
\includegraphics[width=0.5\textwidth]{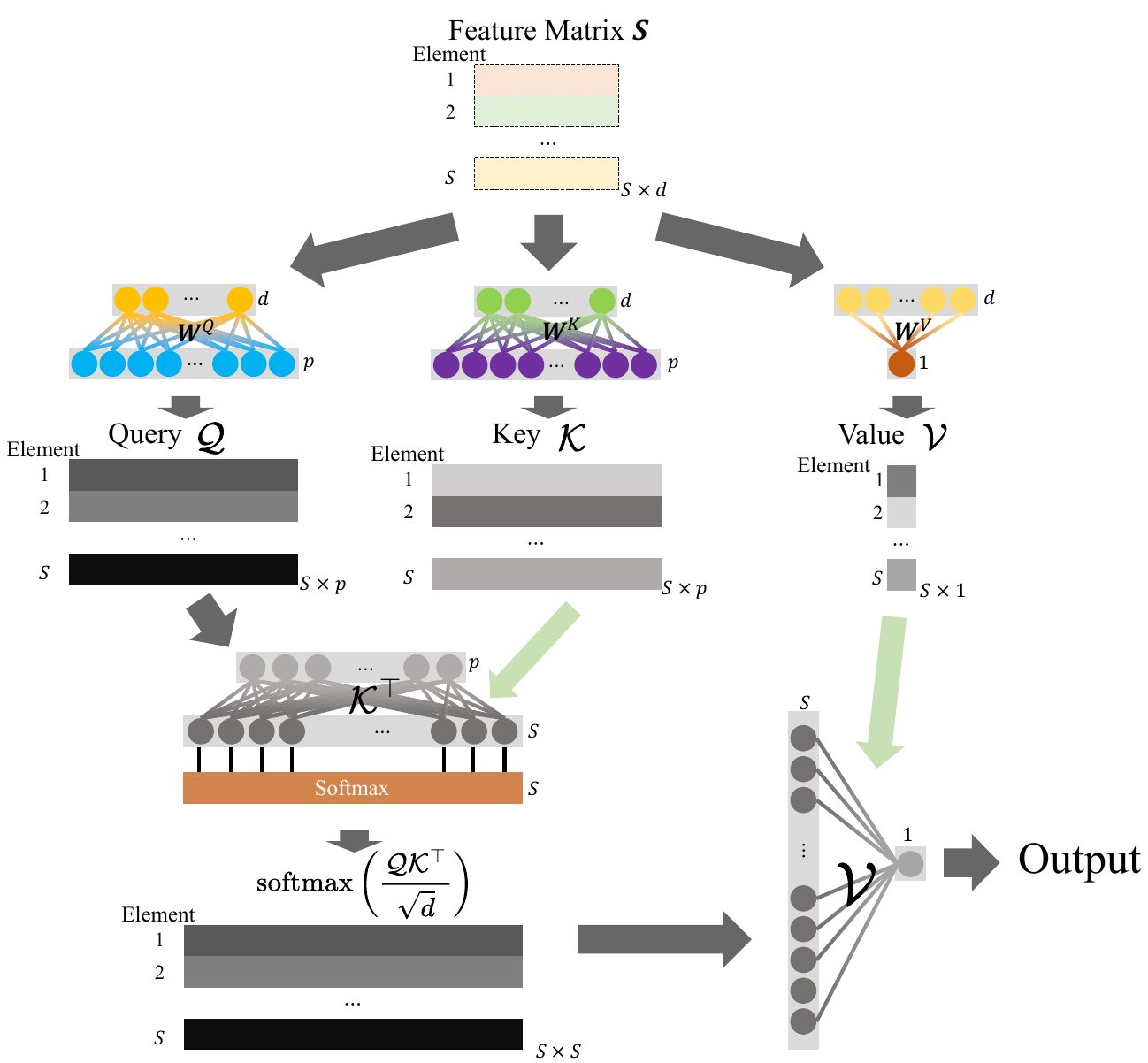}
\vskip -0.1in
\caption{\footnotesize An illustration of the workflow for an attention neural network with an input sequence $\S$.}
\label{fig:self_attention_1}
\vskip -.2in
\end{wrapfigure}
Let the column vector $\theta\in\mbR^m$ denote all trainable weights in a flattened format, with $m$ representing the total number of parameters in the ANN. Figure~\ref{fig:self_attention_1} illustrates the workflow of the ANN. Given a training set of size $N$, $\{(\S_i,\y_i)|\S_i\in\mcS,\y_i\in\mcY\}_N$, the parameters are updated via gradient descent, as shown below:\footnote{Training sequences generally have the same length, corresponding to the maximum length, which is ensured by padding or truncating \citep{yu2023bag,ding2024fewer}. Therefore, this paper focuses on sequences of the same length unless noted otherwise. Results for varying lengths can be directly obtained.}
\begin{equation}
	\theta^{t+1}\gets\theta^t-\frac{\eta}{NS}\sum\ind{i}{1}{N}\sum\ind{j}{1}{S}\nabla_{\theta}\mcL({f_{\theta^t}(\S_i)}_{(j,:)},{\y_i}_{(j,:)}),
\end{equation}
where ${f_{\theta^t}(\S_i)}_{(j,:)}$ refers to the $j$-th row of the output $f_{\theta^t}(\S_i)$, corresponding to the $j$-th element of the input sequence, and ${\y_i}_{(j,:)}$ denotes its associated property value.
Since the learning rate $\eta$ is small enough, the updates remain minimal over multiple iterations, allowing them to be treated as a time derivative and thus expressed as a differential equation~\citep{jacot2018neural,yang2019wide,hron2020infinite}:
\begin{eqnarray}\label{eq:paravar}
	\frac{\partial \theta^t}{\partial t}=-\frac{\eta}{NS}\left[\frac{\partial \mcL(f_{\theta^t}(\S_1),\y_1)}{\partial f_{\theta^t}(\S_1)},\dots,\frac{\partial \mcL(f_{\theta^t}(\S_N),\y_N)}{\partial f_{\theta^t}(\S_N)}\right]\cdot \left[\frac{\partial f_{\theta^t}(\S_i)}{\partial \theta^t}\right]_N~.
\end{eqnarray}

The term $\frac{\partial f_\theta(\S)}{\partial \theta}$ (with the indexes $i$ and $t$ omitted for simplicity), which defines the direction for parameter updates, can be more explicitly written as
\begin{eqnarray}
    \frac{\partial f_\theta(\S)}{\partial \theta}=\Bigg[\underbrace{\frac{\partial f_\theta(\S)}{\partial \W^{V}_{(:,1)}},\dots,\frac{\partial f_\theta(\S)}{\partial \W^{V}_{(:,v)}}}_{\text{w.r.t. the value weight matrix}},\underbrace{\frac{\partial f_\theta(\S)}{\partial \W^{Q}_{(:,1)}},\dots,\frac{\partial f_\theta(\S)}{\partial \W^{Q}_{(:,p)}}}_{\text{w.r.t. the query weight matrix}},\underbrace{\frac{\partial f_\theta(\S)}{\partial \W^{K}_{(:,1)}},\dots,\frac{\partial f_\theta(\S)}{\partial \W^{K}_{(:,p)}}}_{\text{w.r.t. the key weight matrix}}\Bigg]~.
\end{eqnarray}
Here, each term represents the derivative of the output $f_\theta(\S)$ w.r.t. the weight column vectors. Unlike derivatives for multilayer perceptron-based learners, where the input is used only once, \ie, the derivative depends on a single use of the input, the attention mechanism invokes the input three times at $\mcQ$, $\mcK$, and $\mcV$ separately, as depicted in Figure~\ref{fig:self_attention_1}. To clearly demonstrate, in an analytical and explicit manner, how these three invocations allow attention to adaptively assign varying different importance to each element in an input sequence within parameter space, we present an example involving the derivative of an ANN with $v=1$, meaning that each component of the output ${f_{\theta^t}(\S)}_{(j,:)}$ is a scalar:
\begin{eqnarray}
    \frac{\partial f_\theta(\S)}{\partial \theta}=\left[\frac{\partial f_\theta(\S)}{\partial \W^{V}},\frac{\partial f_\theta(\S)}{\partial \W^{Q}_{(:,1)}},\dots,\frac{\partial f_\theta(\S)}{\partial \W^{Q}_{(:,p)}},\frac{\partial f_\theta(\S)}{\partial \W^{K}_{(:,1)}},\dots,\frac{\partial f_\theta(\S)}{\partial \W^{K}_{(:,p)}}\right],
\end{eqnarray}
where the term $\frac{\partial f_\theta(\S)}{\partial \W^{V}}$ has a shape of $S\times d$, and is given by
\begin{eqnarray}\label{eq:grad1}
\frac{\partial f_\theta(\S)}{\partial \W^{V}}=\left[\frac{\exp\left(\mcQ_{(i,:)}\mcK\trsp/\sqrt{d}\right)}{\1\trsp \exp\left(\mcQ_{(i,:)}\mcK\trsp/\sqrt{d}\right)}\right]_S\S,
\end{eqnarray}
where $\exp(\cdot)$ denotes the element-wise exponential operator. For simplicity, we omit the arguments of $\mcQ,\mcK,\mcV$. For $i\in\mbN_{p}$, the term $\frac{\partial f_\theta(\S)}{\partial \W^{Q}_{(:,i)}}$ and $\frac{\partial f_\theta(\S)}{\partial \W^{K}_{(:,i)}}$ are
\begin{eqnarray}\label{eq:grad2}
    \resizebox{.98\hsize}{!}{$\frac{\partial f_\theta(\S)}{\partial \W^{Q}_{(:,i)}}=\bigg[d^{-1/2}\underbrace{\S_{(j,:)}}_{1\times d}\cdot\underbrace{\Bigg(\overbrace{{\mcK_{(:,i)}}\trsp}^{1\times S}\overbrace{\diag\Big(\mathrm{softmax}\big(\mcQ_{(j,i)}{\mcK_{(:,i)}}\trsp/\sqrt{d}\big)\Big)}^{S\times S}\overbrace{\mcV}^{S\times 1}-\overbrace{{\mcK_{(:,i)}}\trsp}^{1\times S}\overbrace{\big(\mathrm{softmax}\big(\mcQ_{(j,i)}{\mcK_{(:,i)}}\trsp/\sqrt{d}\big)\big)\trsp\mathrm{softmax}\big(\mcQ_{(j,i)}{\mcK_{(:,i)}}\trsp/\sqrt{d}\big)}^{S\times S}\overbrace{\mcV}^{S\times 1}\Bigg)}_{\coloneqq\omega_j, 1\times 1}\bigg]_{S\times d}$},
\end{eqnarray}
\begin{eqnarray}\label{eq:grad3}
    \resizebox{.98\hsize}{!}{$\frac{\partial f_\theta(\S)}{\partial \W^{K}_{(:,i)}}=\bigg[d^{-1/2}\underbrace{\S_{(j,:)}}_{1\times d}\cdot\underbrace{\Bigg(\overbrace{{\mcQ_{(:,i)}}\trsp}^{1\times S}\overbrace{\diag\Big(\mathrm{softmax}\big(\mcQ_{(j,i)}{\mcK_{(:,i)}}\trsp/\sqrt{d}\big)\Big)}^{S\times S}\overbrace{\mcV}^{S\times 1}-\overbrace{{\mcQ_{(:,i)}}\trsp}^{1\times S}\overbrace{\big(\mathrm{softmax}\big(\mcQ_{(j,i)}{\mcK_{(:,i)}}\trsp/\sqrt{d}\big)\big)\trsp\mathrm{softmax}\big(\mcQ_{(j,i)}{\mcK_{(:,i)}}\trsp/\sqrt{d}\big)}^{S\times S}\overbrace{\mcV}^{S\times 1}\Bigg)}_{1\times 1}\bigg]_{S\times d}$}~.
\end{eqnarray}
The derivation is provided in Appendix~\ref{app:deri}. For the sake of brevity, we focus on the query gradient, with similar results holding for the key and value gradients. As a result of invoking the input three times, Equation \ref{eq:grad2} reveals that the ANN gradient depends not only on the features of the sequence elements, \ie, $\S_{(j,:)}$, but also on a scalar $\omega_j$ that is specific to each element. 

Specifically, Equation \ref{eq:grad2} explicitly shows that the gradient row order follows the order of elements in the input sequence, meaning the gradient is equivariant w.r.t. reordering the elements. This is in contrast to the gradient in recurrent neural networks \citep{elman1990finding, jordan1997serial}, where the order of the elements determines the power of the recurrent weights. Moreover, this gradient property is derived during the training stage, yet, interestingly, it aligns with the permutation invariance property of self-attention during inference \citep{lee2019set}. 

The scalar $\omega_j$ in Equation \ref{eq:grad2} is computed from $\mcQ$, $\mcK$, and $\mcV$, which reflects the three invocations of input by the attention. It is clear that it is closely associated with the $j$-th element, meaning it is element-specific. This scalar is the importance value that attention assigns to each element, leading to an importance-adaptive update in the parameter space. When all importance values are set to 1, the gradient of the ANN reduces to the derivative of a multilayer perceptron without nonlinear activations and with batch input. Additionally, the explicit expressions in Equations~\ref{eq:paravar}, \ref{eq:grad2}, and \ref{eq:grad3} show that the ANN gradient does not depend on the input sequence length (\ie, the number of elements), as this is averaged out. Instead, it depends on the feature dimension. In other words, the parameter gradient remains unchanged even if the input sequence length $S$ is scaled.

\subsection{The Functional Evolution of ANN}
The importance-adaptive update in the parameter space drives the functional evolution of $f_{\theta}\in\mcH$. This variation in $f_{\theta}$, reflecting how $f_{\theta}$ responds to updates in $\theta$, can be derived using Taylor's theorem as follows:
\begin{eqnarray}
	f(\theta^{t+1})-f(\theta^t)=\langle\nabla_{\theta}f(\theta^t),\theta^{t+1}-\theta^t\rangle+o(\theta^{t+1}-\theta^t),
\end{eqnarray}
where $f(\theta^\dagger)\equiv f_{\theta^\dagger}$. In a manner analogous to the transformation of parameter updates into their differential form, this can also be expressed in a differential form~\citep{zhang2024nonparametric}:
\begin{eqnarray} \label{fparat}
	\frac{\partial f_{\theta^t}}{\partial t}= \underbrace{\left\langle\frac{\partial f(\theta^t)}{\partial \theta^t},\frac{\partial \theta^t}{\partial t}\right\rangle}_{(*)} + o\left(\frac{\partial \theta^t}{\partial t}\right)~.
\end{eqnarray}
By substituting the specific parameter updates, \ie, Equation~\ref{eq:paravar}, into the first-order approximation term $(*)$ of this variation, we obtain
\begin{eqnarray} \label{bfparat}
\frac{\partial f_{\theta^t}}{\partial t}=-\frac{\eta}{NS}\left[\frac{\partial \mcL(f_{\theta^t}(\S_1),\y_1)}{\partial f_{\theta^t}(\S_1)},\dots,\frac{\partial \mcL(f_{\theta^t}(\S_N),\y_N)}{\partial f_{\theta^t}(\S_N)}\right]\cdot\left[K_{\theta^t}(\S_i,\cdot)\right]_N+ o\left(\frac{\partial \theta^t}{\partial t}\right),
\end{eqnarray}
where the symmetric and positive definite $K_{\theta^t}(\S_i,\cdot)\coloneqq\left\langle\frac{\partial f_{\theta^t}(\S_i)}{\partial \theta^t},\frac{\partial f_{\theta^t}(\cdot)}{\partial \theta^t} \right\rangle$ (for detailed derivations and further discussion, see Appendix~\ref{dpntk}). Due to the inclusion of nonlinear activation functions in $f(\theta)$, the nonlinearity of $f(\theta)$ with respect to $\theta$ results in the remainder $o(\theta^{t+1}-\theta^t)$ being nonzero. In a subtle difference, \citealp{jacot2018neural,yang2019wide,hron2020infinite} apply the chain rule directly, giving less focus to the convexity of $\mathcal{L}$ with respect to $\theta$. As a result, the first-order approximation is derived as the variation, with $K_{\theta}$ being referred to as the Attention Neural Tangent Kernel (ANTK). It has been demonstrated that the ANTK remains constant during training when the ANN width, \ie, $d$, is assumed to be infinite~\citep{hron2020infinite}. However, in practical applications, the ANN width does not need to be infinitely large, prompting us to explore the dynamic ANTK (an example of how the ANTK is computed can be found in Figure~\ref{ntk} in Appendix~\ref{dpntk}).

Consider characterizing the variation of $f_{\theta}\in\mcH$ from a high-level, functional viewpoint~\citep{zhang2024nonparametric,zhang2025nonparametric}. Using functional gradient descent, it can be written as
\begin{eqnarray}
\frac{\partial f_{\theta^t}}{\partial t}=-\eta\mcG(\mathcal{L},f^*;f_{\theta^t},\{\S_i\}_N),
\end{eqnarray}
where the functional gradient is expressed as
\begin{eqnarray}
	\mcG(\mathcal{L},f^*;f_{\theta^t},\{\S_i\}_N)=\frac{1}{NS}\left[\frac{\partial \mcL(f_{\theta^t}(\S_1),\y_1)}{\partial f_{\theta^t}(\S_1)},\dots,\frac{\partial \mcL(f_{\theta^t}(\S_N),\y_N)}{\partial f_{\theta^t}(\S_N)}\right]\left[K({\S_i},\cdot)\right]_N~.
\end{eqnarray}
The asymptotic relationship between ANTK and the importance-adaptive canonical kernel~\citep{cancedda2003word,kiraly2019kernels,zhang2024nonparametric} in the context of functional gradient is presented in Theorem~\ref{ntkconverge} below, with the proof provided in Appendix~\ref{pntkconverge}.
\begin{thm}\label{ntkconverge}
Given a convex loss $\mcL$ and a training set $\{(\S_i,\y_i)|\S_i\in\mcS,\y_i\in\mcY\}_N$, the dynamic ANTK, which is derived from performing gradient descent on the parameters of an ANN, converges pointwise to the importance-adaptive canonical kernel in the dual functional gradient with respect to the input sequences. Specifically, it holds that
	\begin{eqnarray}
		\lim_{t\to\infty}K_{\theta^t}({\S_i},\cdot)=K({\S_i},\cdot), \quad \forall i \in\mbN_N~.
	\end{eqnarray}
\end{thm}
This suggests that ANTK, which includes adaptive importance information, serves as a dynamic substitute for the importance-adaptive canonical kernel in functional gradient descent with sequence inputs, aligning the ANN evolution through parameter gradient descent with that in functional gradient descent~\citep{kuk1995asymptotically,hron2020infinite,geifman2020similarity}. This functional insight bridges the teaching of attention learners (\ie, ANNs) with that of importance-adaptive nonparametric learners, while also facilitating further analysis (\eg, a convex functional $\mcL$ retains its convexity with respect to $f_\theta$ from a functional perspective, but is typically nonconvex when considering $\theta$). 
By utilizing the functional insight and applying the canonical kernel~\citep{dou2021training} instead of ANTK (which should be considered \textit{alongside the remainder}), it facilitates deriving sufficient reduction concerning $\mcL$ in Proposition~\ref{slr}, with the proof deferred to Appendix~\ref{pslr}.
\begin{prop} (Sufficient Loss Reduction) \label{slr}
	Let the convex loss $\mcL$ be Lipschitz smooth with a constant $\tau>0$, and suppose the importance-adaptive canonical kernel is bounded above by a constant $\gamma>0$. If the learning rate $\eta$ satisfies $\eta\leq1/(2\tau\gamma)$, then a sufficient reduction in $\mcL$ is guaranteed, as demonstrated by
	\begin{eqnarray}
		\frac{\partial \mathcal{L}}{\partial t}\leq -\frac{\eta\gamma}{2}\left(\frac{1}{NS}\sum_{i=1}^N\sum_{j=1}^S\frac{\partial \mcL\left({f_{\theta^t}(\S_i)}_{(j,:)},{\y_i}_{(j,:)}\right)}{\partial {f_{\theta^t}(\S_i)}_{(j,:)}}\right)^2~.
	\end{eqnarray}
\end{prop}
This indicates that the variation of $\mcL$ over time is capped by a negative value, meaning it decreases by at least the magnitude of this upper bound as time progresses, ensuring convergence.

\subsection{The AtteNT Algorithm}

Building on the understanding of how attention adaptively assigns varying importance in parameter-based gradient descent, as well as the consistency between teaching an ANN and a nonparametric learner, we introduce the AtteNT algorithm. This algorithm is designed to amplify the steepness of the gradients, thereby improving the learning efficiency of the ANN. By considering the gradient as the sum of projections of $\frac{\partial \mathcal{L}(f_{\theta}, f^*)}{\partial f_{\theta}}$ onto the basis $\{K(\S_i, \cdot)\}_N$, the gradient can be increased simply by maximizing the projection $\frac{\partial \mathcal{L}(f_{\theta}(\S_i), \y_i)}{\partial f_{\theta}(\S_i)}$, thus eliminating the need to compute the norm of the basis $\|K(\S_i, \cdot)\|_\mathcal{H}$~\citep{wright2015coordinate, zhang2024nonparametric}.
This suggests that selecting sequences that either maximize $\left\|\frac{\partial \mathcal{L}(f_{\theta}(\S_i), \y_i)}{\partial f_{\theta}(\S_i)}\right\|_2$ or correspond to the larger components of $\frac{\partial \mathcal{L}(f_{\theta}, f^*)}{\partial f_{\theta}}$ can effectively amplify the gradient, indicating that
\begin{eqnarray}
	{\{\S_i\}_m}^*=\underset{\{\S_i\}_m\subseteq\{\S_i\}_N}{\arg\max}\ \left\|\left[\frac{\partial \mcL(f_{\theta}(\S_i),\y_i)}{\partial f_{\theta}(\S_i)}\right]_m\right\|_\mcF,
\end{eqnarray}
with Frobenius norm $\|\cdot\|_\mcF$. From a functional viewpoint, for a convex loss functional $\mcL$, the norm of its partial derivative w.r.t. $f_\theta$, denoted as $\|\frac{\partial\mcL(f_{\theta})}{\partial f_{\theta}}\|_\mathcal{H}$, is positively correlated with $\|f_\theta-f^*\|_\mathcal{H}$. As $f_\theta$ gets closer to $f^*$, the value of $\|\frac{\partial\mcL(f_{\theta})}{\partial f_{\theta}}\|_\mathcal{H}$ decreases~\citep{boyd2004convex, coleman2012calculus}. This relationship becomes especially prominent when $\mcL$ is strongly convex with a larger convexity constant~\citep{kakade2008generalization,arjevani2016lower}. Building on these insights, the AtteNT algorithm selects sequences by
\begin{eqnarray}\label{intalg}
	{\{\S_i\}_m}^*=\underset{\{\S_i\}_m\subseteq\{\S_i\}_N}{\arg\max}\ \left\|\left[f_{\theta}(\S_i)-f^*(\S_i)\right]_m\right\|_\mcF~.
\end{eqnarray}
The pseudocode is provided in Algorithm~\ref{attent_algo}.

\section{Experiments and Results}

To demonstrate the broad effectiveness of the AtteNT Algorithm, we conducted extensive experiments across diverse domains. Our evaluation covered large language models and computer vision models. In addition, we validated performance under multiple training paradigms, including training from scratch, and fine-tuning, consistently achieving strong results.

\begin{table}[b]
\centering
\vspace{-0.1in}
\caption{AtteNT on NLG tasks. The results are averaged over three runs, with standard deviations included. The GSM8K and MATH datasets share a math fine-tuned model, while HumanEval and MBPP use a code fine-tuned model. MT-Bench utilizes a conversation fine-tuned model. The "Avg. time" represents the average fine-tuning time for the three models.}
\label{table:NLP_results}
\small
\resizebox{1.0\linewidth}{!}{
\begin{tabular}{cccccccc}
\toprule
\textbf{Model} & \textbf{AtteNT} & \textbf{Avg. Time(↓)} & \textbf{GSM8K(↑)} & \textbf{MATH(↑)} & \textbf{HumanEval(↑)} & \textbf{MBPP(↑)} & \textbf{MT-Bench(↑)} \\
\midrule
\multirow{2}{*}{LLaMA 2-7B}
            & w/o & 246±1m & 42.96±0.12 & 5.06±0.16 & 18.35±0.31 & 35.65±0.25 & \textbf{4.58±0.01} \\
            & w   & \textbf{213±2m} & \textbf{43.45±0.55} & \textbf{6.48±0.24} & \textbf{21.80±0.38} & \textbf{37.61±0.42} & 4.49±0.02 \\ 
\midrule
\multirow{2}{*}{Mistral-7B} 
            & w/o & 204±2m & 69.13±0.22 & 20.06±0.20 & 43.42±0.14 & 58.52±0.13 & 5.03±0.05 \\
            & w   & \textbf{180±2m} & \textbf{71.26±0.23} & \textbf{23.12±0.44} & \textbf{46.55±0.25} & \textbf{61.74±0.54} & \textbf{5.32±0.04} \\ 
\midrule
\multirow{2}{*}{Gemma-7B}
            & w/o & 228±2m & 75.23±0.45 & 30.52±0.48 & 53.83±0.27 & 65.69±0.29 & 5.42±0.04 \\
            & w   & \textbf{201±2m} & \textbf{77.74±0.32} & \textbf{31.40±0.36} & \textbf{54.26±0.28} & \textbf{66.28±0.46} & \textbf{5.44±0.08} \\
\bottomrule
\end{tabular}
}
\end{table}

\textbf{LLM Scenario.} We evaluate AtteNT algorithms across a diverse set of natural language generation (NLG) tasks. Specifically, we fine-tune LLaMA 2-7B~\citep{touvron2023llama}, Mistral-7B~\citep{jiang2023mistral}, and Gemma-7B~\citep{team2024gemma} on the MetaMathQA dataset~\citep{yu2023metamath} to benchmark their mathematical reasoning capabilities on GSM8K~\citep{cobbe2021gsm8k} and MATH~\citep{hendrycks2021measuring}. To assess coding proficiency, we further fine-tune these models on CodeFeedback~\citep{zheng2024opencodeinterpreter} and evaluate on HumanEval~\citep{chen2021evaluating} and MBPP~\citep{austin2021program}. For conversational ability, we train on WizardLM-Evol-Instruct~\citep{xu2023wizardlm} and evaluate on MT-Bench~~\citep{zheng2024judging}. All experiments are conducted on standardized subsets to ensure comparable training efficiency and are trained for five epochs.

As shown in Table \ref{table:NLP_results}, AtteNT consistently outperforms standard fine-tuning across all evaluated models and tasks while reducing computational overhead. Specifically, fine-tuning LLaMA, Mistral, and Gemma with AtteNT yields accuracy gains of 1.39, 2.14, and 2.42 on GSM8K, and 1.59, 2.89, and 0.76 on MATH. On coding benchmarks, AtteNT improves performance by 3.66\%, 3.25\%, and 0.29\% on HumanEval, and by 2.08\%, 3.25\%, and 3.31\% on MBPP. We further report average fine-tuning time per model under identical data volumes and epoch settings. Since runtime variation arises primarily from AtteNT’s adaptive data selection, the observed results highlight its efficiency: on average, AtteNT reduces training time by 12.78\%, underscoring its advantage in both performance and resource savings.

\textbf{CV Scenario.} The Multi-Modal MAE~\citep{multimae} is designed to address a diverse range of downstream tasks by employing three specialized encoders, each dedicated to processing a distinct image modality. During pre-training, we explore various selection strategies, including different ratios and intervals, to optimize model configuration. The pretraining process is conducted over 800 epochs.

For unsupervised pre-training, we utilize ImageNetS50~\citep{imagenetS} to evaluate the effectiveness of the AttneNT method in enhancing the performance of downstream tasks under suboptimal conditions. Classification performance is assessed using the validation subset of the original dataset, while semantic segmentation and depth estimation tasks are fine-tuned and evaluated on the NYUv2 dataset~\citep{NYUv2}. Given the absence of a large multi-task dataset with aligned task-specific images~\citep{MuST, multimae, task-robust}, we generate pseudo-labels for ImageNetS50 using Mask2Former~\citep{mask2former}.

As shown in Table \ref{tab:cv}, the AtteNT strategy results in a significant reduction in training time, saving 20.58\% during long-duration training from scratch. Additionally, it consistently improves performance across a wide range of downstream tasks. Notably, the depth estimation task exhibits the largest gain, achieving a 5.1\% improvement. We attribute this improvement to the nature of the depth estimation task, which is independent of image type, thus preventing any disruption in data distribution during the selection process. Our experiments demonstrate the efficacy of AtteNT within the ViT architecture.

\revise{The practical performance gains stem directly from the curriculum effect induced by nonparametric teaching~\citep{bengio2009curriculum,wang2021survey,zhang2023nonparametric,zhang2025nonparametric}, which greedily selects the examples that most advance the learner. This naturally creates a curriculum that focuses training on informative, high-gradient examples and avoids gradient dilution from already-mastered ones.}

\begin{table*}[t]
\centering
\caption{AtteNT across various CV downstream tasks. ImageNetS50 uses 50 categories from ImageNet for classification, evaluated by accuracy. NYUv2(S) is a semantic segmentation task with mIoU as the metric. NYUv2(D) involves depth estimation, evaluated using the $\delta_1$ metric, which measures the percentage of pixels with an error ratio below 1.25~\citep{MuST}.}
    \resizebox{1.0\linewidth}{!}{
    \begin{tabular}{cccccc}
        \toprule
        \textbf{Model}&\textbf{AtteNT}&\textbf{Pretraining Time(↓)}&\textbf{ImageNetS50(↑)}&\textbf{NYUv2(S)(↑)}&\textbf{NYUv2(D)(↑)}\\
        \midrule
        \multirow{2}{*}{Multi-Modal MAE}
         & w/o & 1234m &92.2 &51.9  &52.1 \\
         & w  & \textbf{980m}(-20.58\%) & \textbf{92.3} & \textbf{52.6} & \textbf{57.2} \\
        \bottomrule 
    \end{tabular}
    }
\label{tab:cv}
\end{table*}

\begin{table*}[t]
\centering
\caption{Ablation study of AtteNT pre-training configurations. 
Ratio controls how the fraction of selected samples increases over epochs. Interval denotes how often the subset is re-sampled. Selection specifies the sampling strategy: Random (no difficulty prior), Hard (selects only difficult samples), and Soft (Gumbel-Top-k difficulty-aware sampling). The configuration (Incremental, Incremental, Soft) in the \colorbox{red!20}{red} color row is adopted as our final AtteNT setting, as it simultaneously reduces pre-training time and improves performance on all downstream tasks.}
\resizebox{1.0\linewidth}{!}{
    \begin{tabular}{lllclccc}
        \toprule
        \multicolumn{4}{c}{\bf Pre-training} &  & \multicolumn{3}{c}{\bf Downstream} \\ \cline{1-4} \cline{6-8}
         
        Ratio & Interval & Selection & Training time(↓) & & ImageNetS50(↑) & NYUv2(S)(↑) & NYUv2(D)(↑) \\ \midrule
         
         - & - & - & 1234m &          &92.2 &51.9  &52.1 \\
         Cosine &  Incremental & Random & 966m & &88.6 & 45.3 & 49.6 \\
         Cosine & Incremental & Soft & 995m & & 92.1 & 52.2 & 58.8 \\
         Cosine & Fixed & Soft & 1301m & & \textbf{93.2} & 53.6 & 61.4 \\
         \rowcolor{red!20}
         Incremental & Incremental & Soft & 980m & & 92.3 & 52.6 & 57.2 \\
         Incremental & Fixed & Soft & 1319m & & 92.4 & \textbf{53.7} & \textbf{62.1} \\
         Cosine & Incremental & Hard & 972m & & 91.8 & 49.5 & 57.3 \\
         Cosine & Fixed & Hard & 1285m & & 92.1 & 53.0 & 60.8 \\
         Incremental & Incremental & Hard & \textbf{963m} & & 91.4 & 48.4 & 57.2 \\
         Incremental & Fixed & Hard & 1302m & & 92.5 & 52.7 & 59.5 \\
        \bottomrule 
    \end{tabular}
    }
\label{tab:cv_ablation}
\end{table*}

\label{sec:Ablation}
We further present the ablation study results for AttneNT, focusing on the effects of varying data selection strategies and their impact on downstream tasks. Specifically, we investigate the influence of dynamic changes in data selection ratios and step sizes, following the strategy proposed in~\citep{zhang2023nonparametric}. Additionally, we examine how different selection strategies affect the performance of downstream tasks. The Random strategy involves selecting data without any predefined criteria, while the Hard strategy entails deterministic data selection. The Soft strategy, on the other hand, uses probability-based data selection, derived from loss scores. To implement this, we apply the Gumbel-Top-k selection algorithm~\citep{kool2019stochastic} for sampling without replacement. Our results show that the Soft selection strategy achieves the best performance in downstream tasks, significantly improving the model's robustness during training. A more detailed study of the sample ratio can be found in Appendix~\ref{supp:sample_ratio}, and additional comparison results are provided in Appendix~\ref{supp:comparison}.

\section{Concluding Remarks and Future Work}

This paper introduces AtteNT, a novel paradigm that enhances the learning efficiency of attention learners (\ie, ANNs) through nonparametric teaching theory. Specifically, AtteNT reduces the wallclock time required to learn the implicit mapping from sequences to relevant properties by 13.01\% to 20.58\% while consistently preserving and often enhancing the performance across a diverse set of downstream tasks. Moreover, AtteNT establishes a theoretical connection between the evolution of an ANN via parameter-based gradient descent and that of a function using functional gradient descent in nonparametric teaching. This connection between nonparametric teaching theory and ANN training  expands the potential applications of nonparametric teaching in contexts that involve attention mechanisms.

In future work, it would be interesting to explore other variations of AtteNT for different attention learners, such as graph attention networks~\citep{velivckovic2018graph}. Additionally, investigating its robustness under real-world label noise, building upon recent noise-robust advancements~\citep{wei2024vision,hu2024large}, could yield crucial improvements. Another promising direction is to examine the practical applications of AtteNT in improving the efficiency of data-driven methods~\citep{henaff2020data, touvron2021training, muller2022instant} for attention-related tasks, especially in areas like world models.

\section*{Reproducibility statement}
We have taken substantial steps to promote the reproducibility of our research. Appendix~\ref{app:ad} offers a comprehensive overview of the notation, theoretical background, and key algorithm. All proofs for theorems and propositions can be found in Appendix~\ref{app:dp}. Meanwhile, Appendix~\ref{app:ade} provides a comprehensive description of the experimental setup, including training configurations, hyperparameter choices, algorithmic details, and dataset preprocessing procedures. Codes are available at the following link: \href{https://github.com/chen2hang/AtteNT_NonparametricTeaching}{LINK}.

\section*{Acknowledgments}
This work was supported in part by the Theme-based Research Scheme (TRS) project T45-701/22-R of the Research Grants Council of Hong Kong, and in part by the AVNET-HKU Emerging Microelectronics and Ubiquitous Systems (EMUS) Lab. For computer time, this research used the resources of the Supercomputing Laboratory at KAUST.

\bibliography{main.bib}
\bibliographystyle{iclr2026_conference}


\appendix

\newpage

\begin{appendix}
	
	\thispagestyle{plain}
	\begin{center}
		{\Large \bf Appendix}
	\end{center}
	
\end{appendix}

\appendix
\startcontents[appendices]         
\printcontents[appendices]{}{1}{}
\newpage

\section{Additional Discussions} \label{app:ad}
\subsection{Notation Overview} \label{table:notation}
\begin{table}[h!]
\centering
\caption{\footnotesize Summary of Key Notations.}
\begin{tabular}{p{3cm} p{10cm}} 
\toprule
\textbf{Notation} & \textbf{Description} \\
\midrule
$\S_{S \times d}$ & Matrix containing all feature vectors from the ordered sequence $(\x_1,\dots,\x_S)$, with shape $S \times d$ \\
$[x_{s,j}]_d\trsp$ & $d$-dimensional feature vector for the $s$-th element, with components $x_{s,j}$ \\
$\x$ & Short form for $[x_j]_d$ \\
$\S_{(s,:)}$ & The $s$-th row of $\S$, representing the feature vector for the $s$-th element \\
$\S_{(:,i)}$ & The $i$-th column of $\S$, which represents the $i$-th feature across all elements) \\
$\e_i$ & The $i$-th basis vector, having a value of 1 at the $i$-th position and 0 elsewhere \\
$\mcS$ & Collection of all sequences \\
$\y$ & Property associated with the sequences, which can be scalar or vector \\
$\mcY$ & Space of sequential properties, represented as $\mathbb{R}$ or $\mathbb{R}^n$ \\
$\{a_i\}_m$ & A set containing $m$ items \\
$\diag(a_1, \dots, a_m)$ & Diagonal matrix with diagonal entries $a_1, \dots, a_m$ \\
$\diag(a; m)$ & Diagonal matrix with $m$ repeated entries of $a$ \\
$\mbN_S \coloneqq \{1, \dots, S\}$ & Set of natural numbers from $1$ to $S$ \\
$K(\S,\S')$ & A symmetric and positive definite sequence kernel \\
$\mcH$ & Reproducing kernel Hilbert space (RKHS) defined by $K$ \\
$f^*$ & Target mapping from $\mcS$ to $\mcY$ \\
$\y_\dagger$ & Property $f^*(\S_\dagger)$ corresponding to the sequence $\S_\dagger$ \\
\bottomrule
\end{tabular}
\end{table}

\subsection{Functional Gradient} \label{app:fg}

\citealp{zhang2023nonparametric, zhang2023mint} present the chain rule for functional gradients, which is detailed in Lemma~\ref{cr}~\citep{gelfand2000calculus}, and utilize the Fréchet derivative to calculate the derivative of the evaluation functional in RKHS, as shown in Lemma~\ref{ef}~\citep{coleman2012calculus}.

\begin{lem}(Chain rule for functional gradients)
	\label{cr}
	For differentiable functions $G(F): \mbR\mapsto\mbR$ that depend on functionals $F(f):\mcH\mapsto\mbR$, the chain rule is given by 
	\begin{equation}
		\nabla_f G(F(f))=\frac{\partial G(F(f))}{\partial F(f)}\cdot 	\nabla_f F(f)~.
	\end{equation}
\end{lem}
\begin{lem}
	\label{ef}
	The gradient of the evaluation functional at the feature $\x$, denoted as $\left.E_{\x}(f) = f(\x): \mcH \to \mbR\right.$, is given by $\nabla_f E_{\x}(f) = K(\x, \cdot)$, where $K(\x, \x') : \mbR^d \times \mbR^d \to \mbR$ represents a feature-based kernel.
\end{lem}

\subsection{The Derivation of Importance-adaptive Updates in the Parameter Space.} \label{app:deri}

\begin{figure}[h]
    \centering
    \subfloat[\footnotesize Self-attention.]{
        \includegraphics[width=0.33\textwidth]{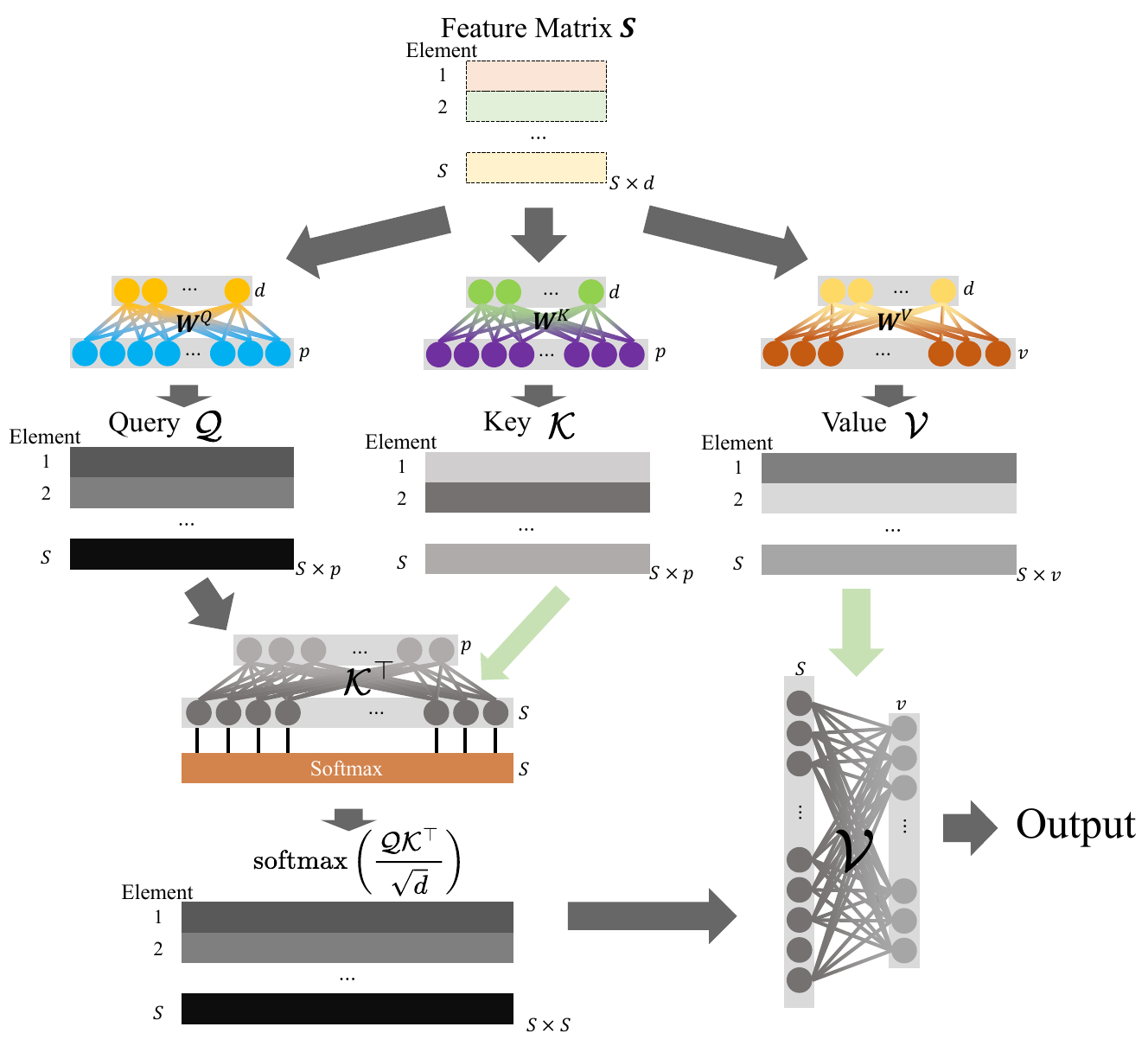}
        \label{fig:self_attention_m}
    }
    \subfloat[\footnotesize Masked self-attention.]{
        \includegraphics[width=0.33\textwidth]{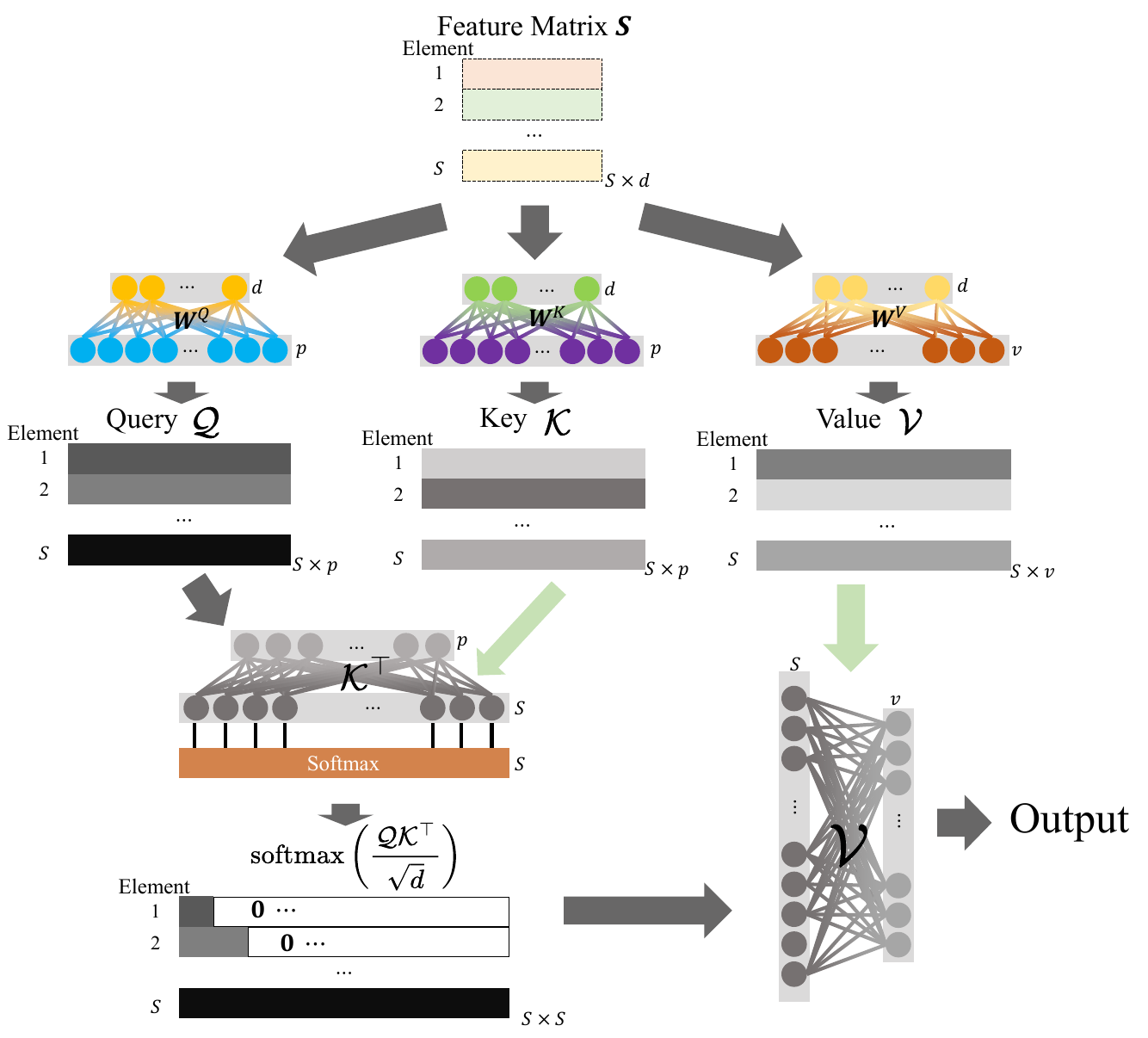}
        \label{fig:masked_self_attention_m}
    }
    \subfloat[Cross-attention.]{
        \includegraphics[width=0.33\textwidth]{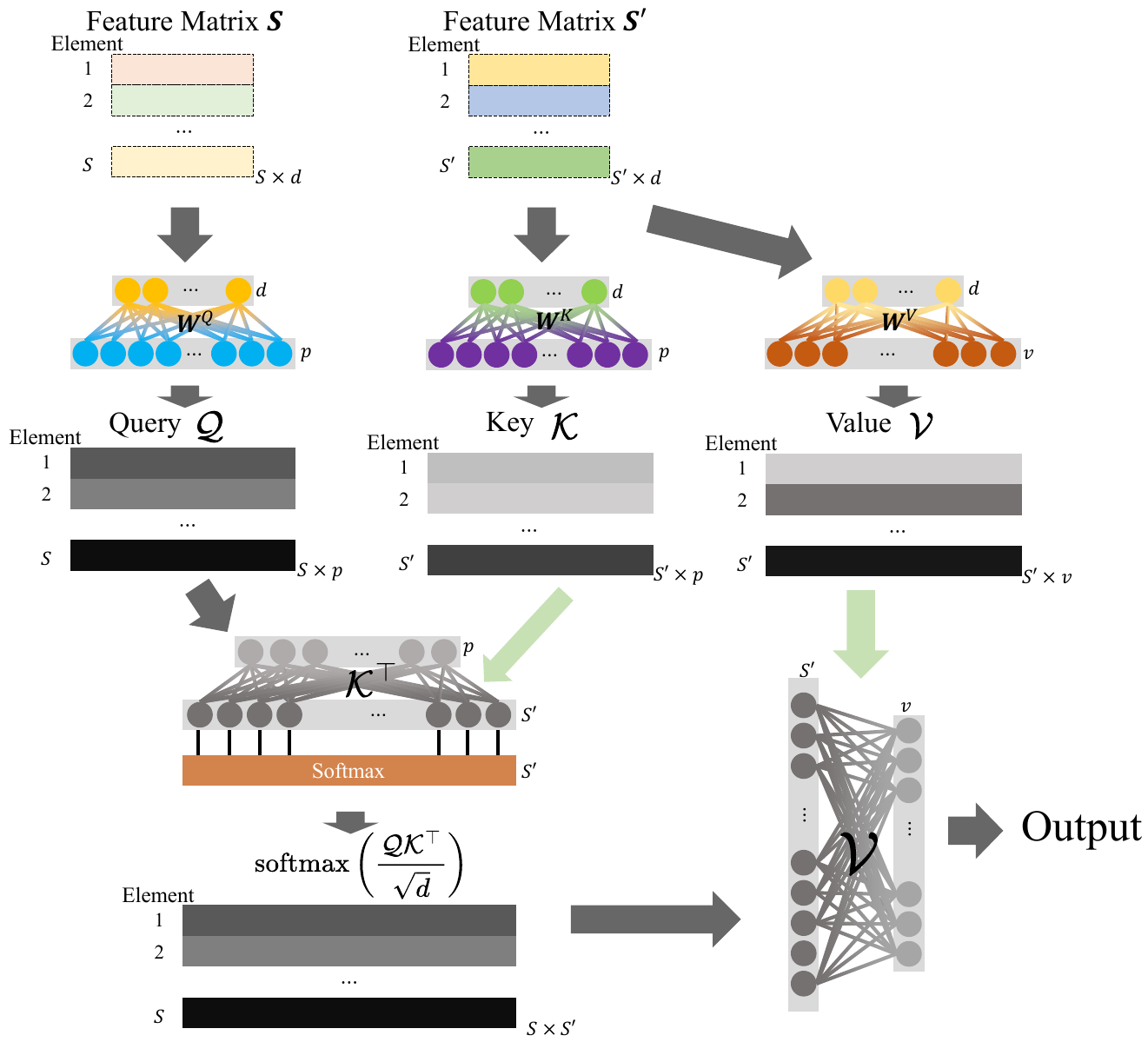}
        \label{fig:cross_attention_m}
    }
    \vskip -0.05in
    \caption{\footnotesize An illustration of the workflow for different multi-output attention learners, with input sequence $\S$ and $\S'$ (in the case of cross-attention).}
    \vskip -0.05in
\end{figure}

Before providing the detailed derivation, we begin by showing visualizations of general single-head attention learners. Figure~\ref{fig:self_attention_m} depicts a multi-output self-attention learner, Figure~\ref{fig:masked_self_attention_m} presents a multi-output masked self-attention learner, and Figure~\ref{fig:cross_attention_m} illustrates a multi-output cross-attention learner. The formulations for the masked self-attention and cross-attention learners are presented in Equation~\ref{eq:general_ann}.
\begin{eqnarray}\label{eq:general_ann}
\text{Masked Self-Attention: }&f_\theta(\S)=\mathrm{softmax}\left(\frac{\mcQ(\S)\mcK(\S)\trsp}{\sqrt{d}}+\M\right)\mcV(\S)\nonumber\\
\text{Cross-Attention: }&f_\theta(\S,\S')=\mathrm{softmax}\left(\frac{\mcQ(\S)\mcK(\S')\trsp}{\sqrt{d}}\right)\mcV(\S'),
\end{eqnarray}
where $\M\in\mbR^{S\times S}$ is a is a strictly upper triangular matrix, with zeros on and below the diagonal and $-\infty$ in every element above the diagonal.

Consider the derivative of an ANN with $v=1$, meaning that each component of the output ${f_{\theta^t}(\S)}_{(j,:)}$ is a scalar:
\begin{eqnarray}
    \frac{\partial f_\theta(\S)}{\partial \theta}&=&\left[\frac{\partial f_\theta(\S)}{\partial \W^{V}},\frac{\partial f_\theta(\S)}{\partial \W^{Q}_{(:,1)}},\dots,\frac{\partial f_\theta(\S)}{\partial \W^{Q}_{(:,p)}},\frac{\partial f_\theta(\S)}{\partial \W^{K}_{(:,1)}},\dots,\frac{\partial f_\theta(\S)}{\partial \W^{K}_{(:,p)}}\right].
\end{eqnarray}

By applying the chain rule, we can compute the derivative of $f_\theta(\S)$ with respect to the weight $\W^{V}$ in the value matrix $\mcV(\S)$.
\begin{eqnarray}\label{eq:desl}
    \frac{\partial f_\theta(\S)}{\partial \W^{V}}&=&\frac{\partial\, \mathrm{softmax}\left(\frac{\mcQ(\S)\mcK(\S)\trsp}{\sqrt{d}}\right)\mcV(\S)}{\partial \W^{V}}\nonumber\\
    &=&\frac{\partial\, \mathrm{softmax}\left(\frac{\S\W^{Q}{\W^{K}}\trsp\S\trsp}{\sqrt{d}}\right)\S\W^{V}}{\partial \W^{V}}\nonumber\\
    &=&\mathrm{softmax}\left(\frac{\S\W^{Q}{\W^{K}}\trsp\S\trsp}{\sqrt{d}}\right)\S\nonumber\\
    &=&\mathrm{softmax}\left(\frac{\mcQ\mcK\trsp}{\sqrt{d}}\right)\S\nonumber\\
    &=&\left[\frac{\exp\left(\mcQ_{(i,:)}\mcK\trsp/\sqrt{d}\right)}{\1\trsp \exp\left(\mcQ_{(i,:)}\mcK\trsp/\sqrt{d}\right)}\right]_S\S,
\end{eqnarray}
where $\exp(\cdot)$ denotes the row-wise exponential operator. The case of $v\geq 2$ represents a multi-dimensional extension, which involves more complex notation but can be derived in a similar manner.

The derivative of $f_\theta(\S)$ with respect to the query weight matrix is more intricate. For $i\in\mbN_{p}$,
\begin{eqnarray}\label{eq:deri_w^q}
    \frac{\partial f_\theta(\S)}{\partial \W^{Q}_{(:,i)}}&=&\frac{\partial\, \mathrm{softmax}\left(\frac{\mcQ(\S)\mcK(\S)\trsp}{\sqrt{d}}\right)\mcV(\S)}{\partial \W^{Q}_{(:,i)}}\nonumber\\
    &=&\frac{\partial\, \mathrm{softmax}\left(\frac{\S\W^{Q}{\W^{K}}\trsp\S\trsp}{\sqrt{d}}\right)\S\W^{V}}{\partial \W^{Q}_{(:,i)}}\nonumber\\
    &=&\frac{\partial\, \mathrm{softmax}\left(\frac{\S\W^{Q}\e_i\e_i\trsp{\W^{K}}\trsp\S\trsp}{\sqrt{d}}\right)\S\W^{V}}{\partial \W^{Q}_{(:,i)}}\nonumber\\
    &=&\left[\begin{array}{c}
		\frac{\partial\, \mathrm{softmax}\left(\frac{\S_{(1,:)}\W^{Q}_{(:,i)}{\W^{K}_{(:,i)}}\trsp\S\trsp}{\sqrt{d}}\right)\S\W^{V}}{\partial \W^{Q}_{(:,i)}}\\ \dots \\\frac{\partial\, \mathrm{softmax}\left(\frac{\S_{(S,:)}\W^{Q}_{(:,i)}{\W^{K}_{(:,i)}}\trsp\S\trsp}{\sqrt{d}}\right)\S\W^{V}}{\partial \W^{Q}_{(:,i)}} \\
	\end{array}\right]\nonumber\\
    &=&\left[\begin{array}{c}
		\frac{\partial \frac{\S_{(1,:)}\W^{Q}_{(:,i)}{\W^{K}_{(:,i)}}\trsp\S\trsp}{\sqrt{d}}}{\partial\W^{Q}_{(:,i)}}\frac{\partial\, \mathrm{softmax}\left(\frac{\S_{(1,:)}\W^{Q}_{(:,i)}{\W^{K}_{(:,i)}}\trsp\S\trsp}{\sqrt{d}}\right)}{\partial \frac{\S_{(1,:)}\W^{Q}_{(:,i)}{\W^{K}_{(:,i)}}\trsp\S\trsp}{\sqrt{d}}}\S\W^{V}\\ \dots \\\frac{\partial \frac{\S_{(S,:)}\W^{Q}_{(:,i)}{\W^{K}_{(:,i)}}\trsp\S\trsp}{\sqrt{d}}}{\partial\W^{Q}_{(:,i)}}\frac{\partial\, \mathrm{softmax}\left(\frac{\S_{(S,:)}\W^{Q}_{(:,i)}{\W^{K}_{(:,i)}}\trsp\S\trsp}{\sqrt{d}}\right)}{\partial \frac{\S_{(S,:)}\W^{Q}_{(:,i)}{\W^{K}_{(:,i)}}\trsp\S\trsp}{\sqrt{d}}}\S\W^{V} \\
	\end{array}\right]~.
\end{eqnarray}
Let’s examine this row by row. For the $j$-th row ($j\in\mbN_S$) of Equation \ref{eq:deri_w^q}, it expressed as:
\begin{eqnarray}\label{eq:deri_w^q_single}
&&\frac{\partial \frac{\S_{(j,:)}\W^{Q}_{(:,i)}{\W^{K}_{(:,i)}}\trsp\S\trsp}{\sqrt{d}}}{\partial\W^{Q}_{(:,i)}}\frac{\partial\, \mathrm{softmax}\left(\frac{\S_{(j,:)}\W^{Q}_{(:,i)}{\W^{K}_{(:,i)}}\trsp\S\trsp}{\sqrt{d}}\right)}{\partial \frac{\S_{(j,:)}\W^{Q}_{(:,i)}{\W^{K}_{(:,i)}}\trsp\S\trsp}{\sqrt{d}}}\S\W^{V}\nonumber\\
&=&\S_{(j,:)}\cdot\frac{1}{\sqrt{d}}{\W^{K}_{(:,i)}}\trsp\S\trsp\frac{\partial\, \mathrm{softmax}\left(\frac{\S_{(j,:)}\W^{Q}_{(:,i)}{\W^{K}_{(:,i)}}\trsp\S\trsp}{\sqrt{d}}\right)}{\partial \frac{\S_{(j,:)}\W^{Q}_{(:,i)}{\W^{K}_{(:,i)}}\trsp\S\trsp}{\sqrt{d}}}\S\W^{V}\nonumber\\
&=&\S_{(j,:)}\cdot\frac{1}{\sqrt{d}}{\W^{K}_{(:,i)}}\trsp\S\trsp\frac{\partial\left( \frac{\exp\left(d^{-1/2}\S_{(j,:)}\W^{Q}_{(:,i)}{\W^{K}_{(:,i)}}\trsp\S\trsp\right)}{\1\trsp \exp\left(d^{-1/2}\S_{(j,:)}\W^{Q}_{(:,i)}{\W^{K}_{(:,i)}}\trsp\S\trsp\right)}\right)}{\partial \, d^{-1/2}\S_{(j,:)}\W^{Q}_{(:,i)}{\W^{K}_{(:,i)}}\trsp\S\trsp}\S\W^{V}\nonumber\\
&=&\S_{(j,:)}\cdot\Bigg(\frac{1}{\sqrt{d}}{\W^{K}_{(:,i)}}\trsp\S\trsp\diag\Big(\mathrm{softmax}\big(\S_{(j,:)}\W^{Q}_{(:,i)}{\W^{K}_{(:,i)}}\trsp\S\trsp/\sqrt{d}\big)\Big)\S\W^{V}\nonumber\\
&&\qquad-\frac{1}{\sqrt{d}}{\W^{K}_{(:,i)}}\trsp\S\trsp\Big(\mathrm{softmax}\big(\S_{(j,:)}\W^{Q}_{(:,i)}{\W^{K}_{(:,i)}}\trsp\S\trsp/\sqrt{d}\big)\Big)\trsp\mathrm{softmax}\big(\S_{(j,:)}\W^{Q}_{(:,i)}{\W^{K}_{(:,i)}}\trsp\S\trsp/\sqrt{d}\big)\S\W^{V}\Bigg)~.\nonumber \\
\end{eqnarray}
The right-hand side of Equation \ref{eq:deri_w^q_single} can be rewritten as
\begin{eqnarray} \label{eq:deri_w^q_single2}
&&\underbrace{\S_{(j,:)}}_{\text{size: }1\times d}\cdot\Bigg(\underbrace{d^{-1/2}}_{1\times 1}\underbrace{{\W^{K}_{(:,i)}}\trsp}_{1\times d}\overbrace{\S\trsp}^{d\times S}\underbrace{\diag\Big(\mathrm{softmax}\big(\overbrace{\S_{(j,:)}}^{1\times d}\overbrace{\W^{Q}_{(:,i)}}^{d\times 1}\overbrace{{\W^{K}_{(:,i)}}\trsp}^{1\times d}\overbrace{\S\trsp}^{d\times S}/\sqrt{d}\big)\Big)}_{S\times S}\underbrace{\S}_{S\times d}\underbrace{\W^{V}}_{d\times 1}\nonumber\\
&&\resizebox{.95\hsize}{!}{$-\underbrace{d^{-1/2}}_{1\times 1}\underbrace{{\W^{K}_{(:,i)}}\trsp}_{1\times d}\overbrace{\S\trsp}^{d\times S}\underbrace{\Big(\mathrm{softmax}\big(\overbrace{\S_{(j,:)}}^{1\times d}\overbrace{\W^{Q}_{(:,i)}}^{d\times 1}\overbrace{{\W^{K}_{(:,i)}}\trsp}^{1\times d}\overbrace{\S\trsp}^{d\times S}/\sqrt{d}\big)\Big)\trsp}_{S\times 1}\underbrace{\mathrm{softmax}\big(\overbrace{\S_{(j,:)}}^{1\times d}\overbrace{\W^{Q}_{(:,i)}}^{d\times 1}\overbrace{{\W^{K}_{(:,i)}}\trsp}^{1\times d}\overbrace{\S\trsp}^{d\times S}/\sqrt{d}\big)}_{1\times S}\underbrace{\S}_{S\times d}\underbrace{\W^{V}}_{d\times 1}$}\Bigg)\nonumber\\
&=&\underbrace{\S_{(j,:)}}_{1\times d}\cdot\Bigg(\underbrace{d^{-1/2}}_{1\times 1}\underbrace{{\mcK_{(:,i)}}\trsp}_{1\times S}\underbrace{\diag\Big(\mathrm{softmax}\big(\overbrace{\mcQ_{(j,i)}}^{1\times 1}\overbrace{{\mcK_{(:,i)}}\trsp}^{1\times S}/\sqrt{d}\big)\Big)}_{S\times S}\underbrace{\mcV}_{S\times 1}\nonumber\\
&&\qquad-\underbrace{d^{-1/2}}_{1\times 1}\underbrace{{\mcK_{(:,i)}}\trsp}_{1\times S}\underbrace{\Big(\mathrm{softmax}\big(\overbrace{\mcQ_{(j,i)}}^{1\times 1}\overbrace{{\mcK_{(:,i)}}\trsp}^{1\times S}/\sqrt{d}\big)\Big)\trsp}_{S\times 1}\underbrace{\mathrm{softmax}\big(\overbrace{\mcQ_{(j,i)}}^{1\times 1}\overbrace{{\mcK_{(:,i)}}\trsp}^{1\times S}/\sqrt{d}\big)}_{1\times S}\underbrace{\mcV}_{S\times 1}\Bigg)\nonumber\\
&=&\resizebox{.95\hsize}{!}{$d^{-1/2}\underbrace{\S_{(j,:)}}_{1\times d}\cdot\underbrace{\Bigg(\overbrace{{\mcK_{(:,i)}}\trsp}^{1\times S}\overbrace{\diag\Big(\mathrm{softmax}\big(\mcQ_{(j,i)}{\mcK_{(:,i)}}\trsp/\sqrt{d}\big)\Big)}^{S\times S}\overbrace{\mcV}^{S\times 1}-\overbrace{{\mcK_{(:,i)}}\trsp}^{1\times S}\overbrace{\big(\mathrm{softmax}\big(\mcQ_{(j,i)}{\mcK_{(:,i)}}\trsp/\sqrt{d}\big)\big)\trsp\mathrm{softmax}\big(\mcQ_{(j,i)}{\mcK_{(:,i)}}\trsp/\sqrt{d}\big)}^{S\times S}\overbrace{\mcV}^{S\times 1}\Bigg)}_{1\times 1}$}~.\nonumber\\
\end{eqnarray}
By combining Equation~\ref{eq:deri_w^q}, \ref{eq:deri_w^q_single} and \ref{eq:deri_w^q_single2}, we derive
\begin{eqnarray}\label{eq:deri_w^q_all}
&&\frac{\partial f_\theta(\S)}{\partial \W^{Q}_{(:,i)}}\nonumber\\
&=&\resizebox{.95\hsize}{!}{$\left[\begin{array}{c}
		d^{-1/2}\underbrace{\S_{(1,:)}}_{1\times d}\cdot\underbrace{\Bigg(\overbrace{{\mcK_{(:,i)}}\trsp}^{1\times S}\overbrace{\diag\Big(\mathrm{softmax}\big(\mcQ_{(1,i)}{\mcK_{(:,i)}}\trsp/\sqrt{d}\big)\Big)}^{S\times S}\overbrace{\mcV}^{S\times 1}-\overbrace{{\mcK_{(:,i)}}\trsp}^{1\times S}\overbrace{\big(\mathrm{softmax}\big(\mcQ_{(1,i)}{\mcK_{(:,i)}}\trsp/\sqrt{d}\big)\big)\trsp\mathrm{softmax}\big(\mcQ_{(1,i)}{\mcK_{(:,i)}}\trsp/\sqrt{d}\big)}^{S\times S}\overbrace{\mcV}^{S\times 1}\Bigg)}_{1\times 1} \\
        \dots \\
        d^{-1/2}\underbrace{\S_{(S,:)}}_{1\times d}\cdot\underbrace{\Bigg(\overbrace{{\mcK_{(:,i)}}\trsp}^{1\times S}\overbrace{\diag\Big(\mathrm{softmax}\big(\mcQ_{(S,i)}{\mcK_{(:,i)}}\trsp/\sqrt{d}\big)\Big)}^{S\times S}\overbrace{\mcV}^{S\times 1}-\overbrace{{\mcK_{(:,i)}}\trsp}^{1\times S}\overbrace{\big(\mathrm{softmax}\big(\mcQ_{(S,i)}{\mcK_{(:,i)}}\trsp/\sqrt{d}\big)\big)\trsp\mathrm{softmax}\big(\mcQ_{(S,i)}{\mcK_{(:,i)}}\trsp/\sqrt{d}\big)}^{S\times S}\overbrace{\mcV}^{S\times 1}\Bigg)}_{1\times 1}
	\end{array}\right]_{S\times d}$}\nonumber\\
&=&\resizebox{.95\hsize}{!}{$\left[d^{-1/2}\S_{(j,:)}\cdot\Bigg({\mcK_{(:,i)}}\trsp\diag\Big(\mathrm{softmax}\big(\mcQ_{(j,i)}{\mcK_{(:,i)}}\trsp/\sqrt{d}\big)\Big)\mcV-{\mcK_{(:,i)}}\trsp\big(\mathrm{softmax}\big(\mcQ_{(j,i)}{\mcK_{(:,i)}}\trsp/\sqrt{d}\big)\big)\trsp\mathrm{softmax}\big(\mcQ_{(j,i)}{\mcK_{(:,i)}}\trsp/\sqrt{d}\big)\mcV\Bigg)\right]_{S\times d}$}\nonumber\\
&=&\diag\Bigg(\resizebox{.85\hsize}{!}{${\mcK_{(:,i)}}\trsp\diag\Big(\mathrm{softmax}\big(\mcQ_{(1,i)}{\mcK_{(:,i)}}\trsp/\sqrt{d}\big)\Big)\mcV-{\mcK_{(:,i)}}\trsp\big(\mathrm{softmax}\big(\mcQ_{(1,i)}{\mcK_{(:,i)}}\trsp/\sqrt{d}\big)\big)\trsp\mathrm{softmax}\big(\mcQ_{(1,i)}{\mcK_{(:,i)}}\trsp/\sqrt{d}\big)\mcV$},\nonumber\\
&&\qquad\dots,\resizebox{.85\hsize}{!}{${\mcK_{(:,i)}}\trsp\diag\Big(\mathrm{softmax}\big(\mcQ_{(S,i)}{\mcK_{(:,i)}}\trsp/\sqrt{d}\big)\Big)\mcV-{\mcK_{(:,i)}}\trsp\big(\mathrm{softmax}\big(\mcQ_{(S,i)}{\mcK_{(:,i)}}\trsp/\sqrt{d}\big)\big)\trsp\mathrm{softmax}\big(\mcQ_{(S,i)}{\mcK_{(:,i)}}\trsp/\sqrt{d}\big)\mcV$}\Bigg)\S/\sqrt{d}~.\nonumber\\
\end{eqnarray}

The derivative of $f_\theta(\S)$ with respect to the key weight matrix is derived in a manner similar to that of the query weight matrix. For $i\in\mbN_{p}$,
\begin{eqnarray}\label{eq:deri_w^k}
    &&\frac{\partial f_\theta(\S)}{\partial \W^{K}_{(:,i)}}\nonumber\\
    &=&\frac{\partial\, \mathrm{softmax}\left(\frac{\mcQ(\S)\mcK(\S)\trsp}{\sqrt{d}}\right)\mcV(\S)}{\partial \W^{K}_{(:,i)}}\nonumber\\
    &=&\frac{\partial\, \mathrm{softmax}\left(\frac{\S\W^{Q}{\W^{K}}\trsp\S\trsp}{\sqrt{d}}\right)\S\W^{V}}{\partial \W^{K}_{(:,i)}}\nonumber\\
    &=&\frac{\partial\, \mathrm{softmax}\left(\frac{\S\W^{Q}{\left(\W^{K}\e_i\e_i\trsp\right)}\trsp\S\trsp}{\sqrt{d}}\right)\S\W^{V}}{\partial \W^{K}_{(:,i)}}\nonumber\\
    &=&\left[\begin{array}{c}
		\frac{\partial\, \mathrm{softmax}\left(\frac{\S_{(1,:)}\W^{Q}_{(:,i)}{\W^{K}_{(:,i)}}\trsp\S\trsp}{\sqrt{d}}\right)\S\W^{V}}{\partial \W^{K}_{(:,i)}}\\ \dots \\\frac{\partial\, \mathrm{softmax}\left(\frac{\S_{(S,:)}\W^{Q}_{(:,i)}{\W^{K}_{(:,i)}}\trsp\S\trsp}{\sqrt{d}}\right)\S\W^{V}}{\partial \W^{K}_{(:,i)}} \\
	\end{array}\right]\nonumber\\
    &=&\left[\begin{array}{c}
		\frac{\partial \frac{\S_{(1,:)}\W^{Q}_{(:,i)}{\W^{K}_{(:,i)}}\trsp\S\trsp}{\sqrt{d}}}{\partial\W^{K}_{(:,i)}}\frac{\partial\, \mathrm{softmax}\left(\frac{\S_{(1,:)}\W^{Q}_{(:,i)}{\W^{K}_{(:,i)}}\trsp\S\trsp}{\sqrt{d}}\right)}{\partial \frac{\S_{(1,:)}\W^{Q}_{(:,i)}{\W^{K}_{(:,i)}}\trsp\S\trsp}{\sqrt{d}}}\S\W^{V}\\ \dots \\\frac{\partial \frac{\S_{(S,:)}\W^{Q}_{(:,i)}{\W^{K}_{(:,i)}}\trsp\S\trsp}{\sqrt{d}}}{\partial\W^{K}_{(:,i)}}\frac{\partial\, \mathrm{softmax}\left(\frac{\S_{(S,:)}\W^{Q}_{(:,i)}{\W^{K}_{(:,i)}}\trsp\S\trsp}{\sqrt{d}}\right)}{\partial \frac{\S_{(S,:)}\W^{Q}_{(:,i)}{\W^{K}_{(:,i)}}\trsp\S\trsp}{\sqrt{d}}}\S\W^{V} \\
	\end{array}\right]~.
\end{eqnarray}
Similarly, let’s examine it row by row. For the $j$-th row ($j\in\mbN_S$) of Equation \ref{eq:deri_w^k}, it is
\begin{eqnarray}\label{eq:deri_w^k_single}
&&\frac{\partial \frac{\S_{(j,:)}\W^{Q}_{(:,i)}{\W^{K}_{(:,i)}}\trsp\S\trsp}{\sqrt{d}}}{\partial\W^{K}_{(:,i)}}\frac{\partial\, \mathrm{softmax}\left(\frac{\S_{(j,:)}\W^{Q}_{(:,i)}{\W^{K}_{(:,i)}}\trsp\S\trsp}{\sqrt{d}}\right)}{\partial \frac{\S_{(j,:)}\W^{Q}_{(:,i)}{\W^{K}_{(:,i)}}\trsp\S\trsp}{\sqrt{d}}}\S\W^{V}\nonumber\\
&=&\S_{(j,:)}\cdot\frac{1}{\sqrt{d}}{\W^{Q}_{(:,i)}}\trsp\S\trsp\frac{\partial\, \mathrm{softmax}\left(\frac{\S_{(j,:)}\W^{Q}_{(:,i)}{\W^{K}_{(:,i)}}\trsp\S\trsp}{\sqrt{d}}\right)}{\partial \frac{\S_{(j,:)}\W^{Q}_{(:,i)}{\W^{K}_{(:,i)}}\trsp\S\trsp}{\sqrt{d}}}\S\W^{V}\nonumber\\
&=&\S_{(j,:)}\cdot\frac{1}{\sqrt{d}}{\W^{Q}_{(:,i)}}\trsp\S\trsp\frac{\partial\left( \frac{\exp\left(d^{-1/2}\S_{(j,:)}\W^{Q}_{(:,i)}{\W^{K}_{(:,i)}}\trsp\S\trsp\right)}{\1\trsp \exp\left(d^{-1/2}\S_{(j,:)}\W^{Q}_{(:,i)}{\W^{K}_{(:,i)}}\trsp\S\trsp\right)}\right)}{\partial \, d^{-1/2}\S_{(j,:)}\W^{Q}_{(:,i)}{\W^{K}_{(:,i)}}\trsp\S\trsp}\S\W^{V}\nonumber\\
&=&\S_{(j,:)}\cdot\Bigg(\frac{1}{\sqrt{d}}{\W^{Q}_{(:,i)}}\trsp\S\trsp\diag\Big(\mathrm{softmax}\big(\S_{(j,:)}\W^{Q}_{(:,i)}{\W^{K}_{(:,i)}}\trsp\S\trsp/\sqrt{d}\big)\Big)\S\W^{V}\nonumber\\
&&\qquad-\frac{1}{\sqrt{d}}{\W^{Q}_{(:,i)}}\trsp\S\trsp\Big(\mathrm{softmax}\big(\S_{(j,:)}\W^{Q}_{(:,i)}{\W^{K}_{(:,i)}}\trsp\S\trsp/\sqrt{d}\big)\Big)\trsp\mathrm{softmax}\big(\S_{(j,:)}\W^{Q}_{(:,i)}{\W^{K}_{(:,i)}}\trsp\S\trsp/\sqrt{d}\big)\S\W^{V}\Bigg)~.\nonumber\\
\end{eqnarray}
The right-hand side of Equation \ref{eq:deri_w^k_single} can be simplified to:
\begin{eqnarray} \label{eq:deri_w^k_single2}
&&\underbrace{\S_{(j,:)}}_{\text{size: }1\times d}\cdot\Bigg(\underbrace{d^{-1/2}}_{1\times 1}\underbrace{{\W^{Q}_{(:,i)}}\trsp}_{1\times d}\overbrace{\S\trsp}^{d\times S}\underbrace{\diag\Big(\mathrm{softmax}\big(\overbrace{\S_{(j,:)}}^{1\times d}\overbrace{\W^{Q}_{(:,i)}}^{d\times 1}\overbrace{{\W^{K}_{(:,i)}}\trsp}^{1\times d}\overbrace{\S\trsp}^{d\times S}/\sqrt{d}\big)\Big)}_{S\times S}\underbrace{\S}_{S\times d}\underbrace{\W^{V}}_{d\times 1}\nonumber\\
&&\resizebox{.95\hsize}{!}{$-\underbrace{d^{-1/2}}_{1\times 1}\underbrace{{\W^{Q}_{(:,i)}}\trsp}_{1\times d}\overbrace{\S\trsp}^{d\times S}\underbrace{\Big(\mathrm{softmax}\big(\overbrace{\S_{(j,:)}}^{1\times d}\overbrace{\W^{Q}_{(:,i)}}^{d\times 1}\overbrace{{\W^{K}_{(:,i)}}\trsp}^{1\times d}\overbrace{\S\trsp}^{d\times S}/\sqrt{d}\big)\Big)\trsp}_{S\times 1}\underbrace{\mathrm{softmax}\big(\overbrace{\S_{(j,:)}}^{1\times d}\overbrace{\W^{Q}_{(:,i)}}^{d\times 1}\overbrace{{\W^{K}_{(:,i)}}\trsp}^{1\times d}\overbrace{\S\trsp}^{d\times S}/\sqrt{d}\big)}_{1\times S}\underbrace{\S}_{S\times d}\underbrace{\W^{V}}_{d\times 1}$}\Bigg)\nonumber\\
&=&\underbrace{\S_{(j,:)}}_{1\times d}\cdot\Bigg(\underbrace{d^{-1/2}}_{1\times 1}\underbrace{{\mcQ_{(:,i)}}\trsp}_{1\times S}\underbrace{\diag\Big(\mathrm{softmax}\big(\overbrace{\mcQ_{(j,i)}}^{1\times 1}\overbrace{{\mcK_{(:,i)}}\trsp}^{1\times S}/\sqrt{d}\big)\Big)}_{S\times S}\underbrace{\mcV}_{S\times 1}\nonumber\\
&&\qquad-\underbrace{d^{-1/2}}_{1\times 1}\underbrace{{\mcQ_{(:,i)}}\trsp}_{1\times S}\underbrace{\Big(\mathrm{softmax}\big(\overbrace{\mcQ_{(j,i)}}^{1\times 1}\overbrace{{\mcK_{(:,i)}}\trsp}^{1\times S}/\sqrt{d}\big)\Big)\trsp}_{S\times 1}\underbrace{\mathrm{softmax}\big(\overbrace{\mcQ_{(j,i)}}^{1\times 1}\overbrace{{\mcK_{(:,i)}}\trsp}^{1\times S}/\sqrt{d}\big)}_{1\times S}\underbrace{\mcV}_{S\times 1}\Bigg)\nonumber\\
&=&\resizebox{.95\hsize}{!}{$d^{-1/2}\underbrace{\S_{(j,:)}}_{1\times d}\cdot\underbrace{\Bigg(\overbrace{{\mcQ_{(:,i)}}\trsp}^{1\times S}\overbrace{\diag\Big(\mathrm{softmax}\big(\mcQ_{(j,i)}{\mcK_{(:,i)}}\trsp/\sqrt{d}\big)\Big)}^{S\times S}\overbrace{\mcV}^{S\times 1}-\overbrace{{\mcQ_{(:,i)}}\trsp}^{1\times S}\overbrace{\big(\mathrm{softmax}\big(\mcQ_{(j,i)}{\mcK_{(:,i)}}\trsp/\sqrt{d}\big)\big)\trsp\mathrm{softmax}\big(\mcQ_{(j,i)}{\mcK_{(:,i)}}\trsp/\sqrt{d}\big)}^{S\times S}\overbrace{\mcV}^{S\times 1}\Bigg)}_{1\times 1}$}~.\nonumber\\
\end{eqnarray}
By merging Equation~\ref{eq:deri_w^k}, \ref{eq:deri_w^k_single} and \ref{eq:deri_w^k_single2}, we get:
\begin{eqnarray}\label{eq:deri_w^k_all}
&&\frac{\partial f_\theta(\S)}{\partial \W^{K}_{(:,i)}}\nonumber\\
&=&\resizebox{.95\hsize}{!}{$\left[\begin{array}{c}
		d^{-1/2}\underbrace{\S_{(1,:)}}_{1\times d}\cdot\underbrace{\Bigg(\overbrace{{\mcQ_{(:,i)}}\trsp}^{1\times S}\overbrace{\diag\Big(\mathrm{softmax}\big(\mcQ_{(1,i)}{\mcK_{(:,i)}}\trsp/\sqrt{d}\big)\Big)}^{S\times S}\overbrace{\mcV}^{S\times 1}-\overbrace{{\mcQ_{(:,i)}}\trsp}^{1\times S}\overbrace{\big(\mathrm{softmax}\big(\mcQ_{(1,i)}{\mcK_{(:,i)}}\trsp/\sqrt{d}\big)\big)\trsp\mathrm{softmax}\big(\mcQ_{(1,i)}{\mcK_{(:,i)}}\trsp/\sqrt{d}\big)}^{S\times S}\overbrace{\mcV}^{S\times 1}\Bigg)}_{1\times 1} \\
        \dots \\
        d^{-1/2}\underbrace{\S_{(S,:)}}_{1\times d}\cdot\underbrace{\Bigg(\overbrace{{\mcQ_{(:,i)}}\trsp}^{1\times S}\overbrace{\diag\Big(\mathrm{softmax}\big(\mcQ_{(S,i)}{\mcK_{(:,i)}}\trsp/\sqrt{d}\big)\Big)}^{S\times S}\overbrace{\mcV}^{S\times 1}-\overbrace{{\mcQ_{(:,i)}}\trsp}^{1\times S}\overbrace{\big(\mathrm{softmax}\big(\mcQ_{(S,i)}{\mcK_{(:,i)}}\trsp/\sqrt{d}\big)\big)\trsp\mathrm{softmax}\big(\mcQ_{(S,i)}{\mcK_{(:,i)}}\trsp/\sqrt{d}\big)}^{S\times S}\overbrace{\mcV}^{S\times 1}\Bigg)}_{1\times 1}
	\end{array}\right]_{S\times d}$}\nonumber\\
&=&\resizebox{.95\hsize}{!}{$\left[d^{-1/2}\S_{(j,:)}\cdot\Bigg({\mcQ_{(:,i)}}\trsp\diag\Big(\mathrm{softmax}\big(\mcQ_{(j,i)}{\mcK_{(:,i)}}\trsp/\sqrt{d}\big)\Big)\mcV-{\mcQ_{(:,i)}}\trsp\big(\mathrm{softmax}\big(\mcQ_{(j,i)}{\mcK_{(:,i)}}\trsp/\sqrt{d}\big)\big)\trsp\mathrm{softmax}\big(\mcQ_{(j,i)}{\mcK_{(:,i)}}\trsp/\sqrt{d}\big)\mcV\Bigg)\right]_{S\times d}$}\nonumber\\
&=&\diag\Bigg(\resizebox{.85\hsize}{!}{${\mcQ_{(:,i)}}\trsp\diag\Big(\mathrm{softmax}\big(\mcQ_{(1,i)}{\mcK_{(:,i)}}\trsp/\sqrt{d}\big)\Big)\mcV-{\mcQ_{(:,i)}}\trsp\big(\mathrm{softmax}\big(\mcQ_{(1,i)}{\mcK_{(:,i)}}\trsp/\sqrt{d}\big)\big)\trsp\mathrm{softmax}\big(\mcQ_{(1,i)}{\mcK_{(:,i)}}\trsp/\sqrt{d}\big)\mcV$},\nonumber\\
&&\qquad\dots,\resizebox{.85\hsize}{!}{${\mcQ_{(:,i)}}\trsp\diag\Big(\mathrm{softmax}\big(\mcQ_{(S,i)}{\mcK_{(:,i)}}\trsp/\sqrt{d}\big)\Big)\mcV-{\mcQ_{(:,i)}}\trsp\big(\mathrm{softmax}\big(\mcQ_{(S,i)}{\mcK_{(:,i)}}\trsp/\sqrt{d}\big)\big)\trsp\mathrm{softmax}\big(\mcQ_{(S,i)}{\mcK_{(:,i)}}\trsp/\sqrt{d}\big)\mcV$}\Bigg)\S/\sqrt{d}~.\nonumber\\
\end{eqnarray}

From Equations~\ref{eq:desl}, \ref{eq:deri_w^q_all}, and \ref{eq:deri_w^k_all}, it can be observed that the ANN gradient for a single sequence resembles that for a batch of feature vector inputs, as the gradient can be decomposed for each element. This demonstrates the parallelization-friendly nature of the attention mechanism from a gradient perspective.

These results can be directly extended to the case where each component of the output ${f_{\theta^t}(\S)}_{(j,:)}$ is a vector, by considering a multi-dimensional setting \citep{zhang2023mint}. The extension to multi-head cases can be done by broadcasting, which involves repeating the derivation in parallel as many times as there are heads.

\subsection{Attention Neural Tangent Kernel (ANTK)} \label{dpntk}

\begin{figure}[t]
	\begin{center}
		\centerline{\includegraphics[width=\textwidth]{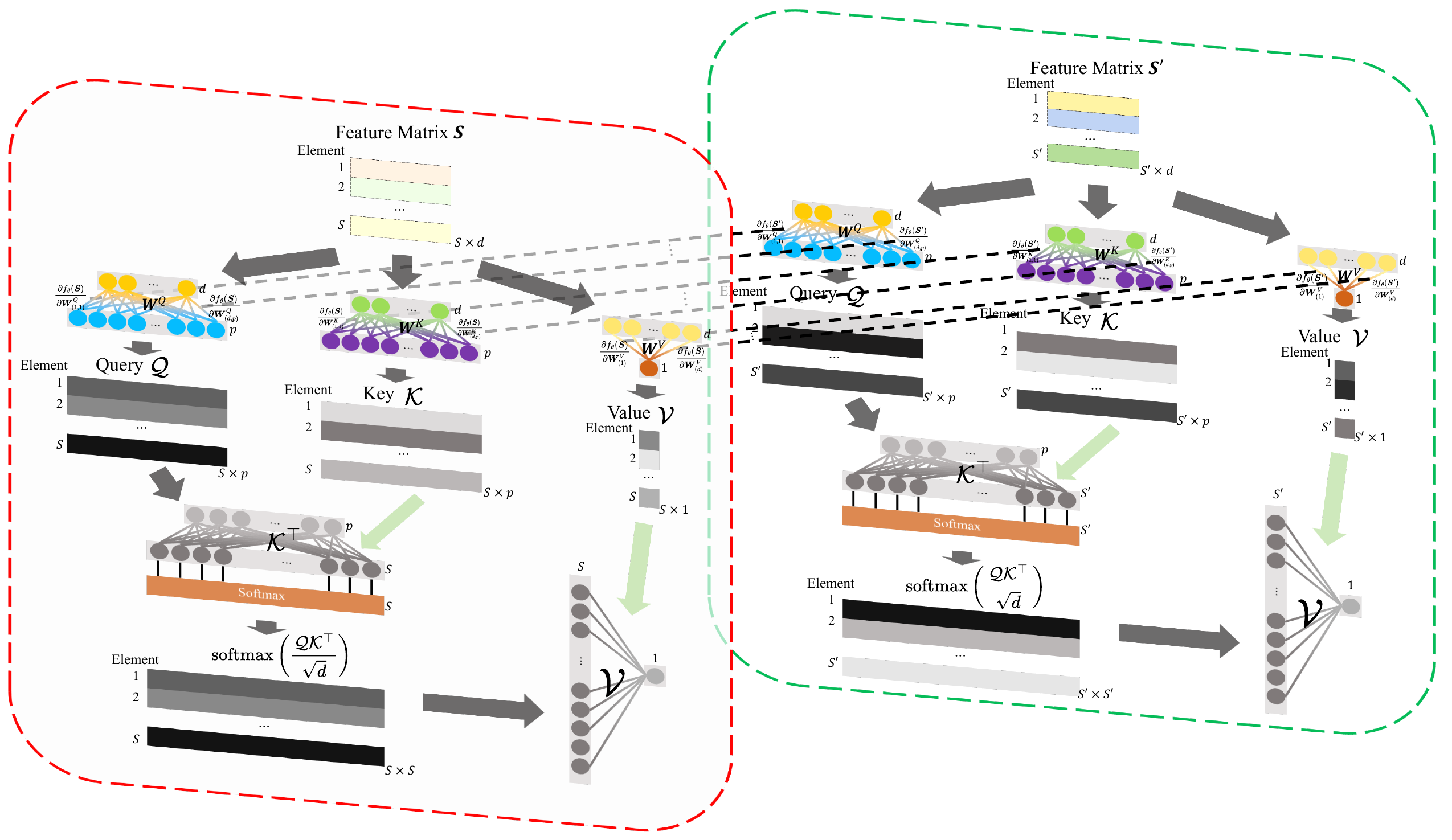}}
		\caption{\footnotesize Graphical depiction of the ANTK computation process: $K_{\theta}(\S_{S},\S'_{S'})=\left\langle\frac{\partial f_{\theta}(\S)}{\partial \theta},\frac{\partial f_{\theta}(\S')}{\partial \theta} \right\rangle=\bigg[\frac{\partial {f_{\theta}(\S)}_{(i,:)}}{\partial \W^{V}_{(1)}}\frac{\partial {f_{\theta}(\S')}_{(j,:)}}{\partial \W^{V}_{(1)}}+\dots+\frac{\partial {f_{\theta}(\S)}_{(i,:)}}{\partial \W^{V}_{(d)}}\frac{\partial {f_{\theta}(\S')}_{(j,:)}}{\partial \W^{V}_{(d)}}+\frac{\partial {f_{\theta}(\S)}_{(i,:)}}{\partial \W^{Q}_{(1,1)}}\frac{\partial {f_{\theta}(\S')}_{(j,:)}}{\partial \W^{Q}_{(1,1)}}+\dots+\frac{\partial {f_{\theta}(\S)}_{(i,:)}}{\partial \W^{Q}_{(d,p)}}\frac{\partial {f_{\theta}(\S')}_{(j,:)}}{\partial \W^{Q}_{(d,p)}}+\frac{\partial {f_{\theta}(\S)}_{(i,:)}}{\partial \W^{K}_{(1,1)}}\frac{\partial {f_{\theta}(\S')}_{(j,:)}}{\partial \W^{K}_{(1,1)}}+\dots+\frac{\partial {f_{\theta}(\S)}_{(i,:)}}{\partial \W^{K}_{(d,p)}}\frac{\partial {f_{\theta}(\S')}_{(j,:)}}{\partial \W^{K}_{(d,p)}}\bigg]_{S\times S';i\in \mbN_S,j\in\mbN_{S'}}$.}
		\label{ntk}
	\end{center}
	\vskip -0.2in
\end{figure}

By incorporating the parameter evolution (\ie, Equation~\ref{eq:paravar})
\begin{eqnarray}
	\frac{\partial \theta^t}{\partial t}=-\frac{\eta}{NS}\left[\frac{\partial \mcL(f_{\theta^t}(\S_1),\y_1)}{\partial f_{\theta^t}(\S_1)},\dots,\frac{\partial \mcL(f_{\theta^t}(\S_N),\y_N)}{\partial f_{\theta^t}(\S_N)}\right]\cdot \left[\frac{\partial f_{\theta^t}(\S_i)}{\partial \theta^t}\right]_N.
\end{eqnarray}
into the first-order approximation term $(*)$ of Equation~\ref{fparat}, we derive
\begin{eqnarray}
	(*)&=&\left\langle\frac{\partial f_{\theta^t}(\cdot)}{\partial \theta^t},-\frac{\eta}{NS}\left[\frac{\partial \mcL(f_{\theta^t}(\S_1),\y_1)}{\partial f_{\theta^t}(\S_1)},\dots,\frac{\partial \mcL(f_{\theta^t}(\S_N),\y_N)}{\partial f_{\theta^t}(\S_N)}\right]\cdot \left[\frac{\partial f_{\theta^t}(\S_i)}{\partial \theta^t}\right]_N\right\rangle\nonumber\\
	&=&-\frac{\eta}{NS}\left[\frac{\partial \mcL(f_{\theta^t}(\S_1),\y_1)}{\partial f_{\theta^t}(\S_1)},\dots,\frac{\partial \mcL(f_{\theta^t}(\S_N),\y_N)}{\partial f_{\theta^t}(\S_N)}\right]\cdot \left\langle\frac{\partial f_{\theta^t}(\cdot)}{\partial \theta^t}, \left[\frac{\partial f_{\theta^t}(\S_i)}{\partial \theta^t}\right]_N\right\rangle\nonumber\\
	&=&-\frac{\eta}{NS}\left[\frac{\partial \mcL(f_{\theta^t}(\S_1),\y_1)}{\partial f_{\theta^t}(\S_1)},\dots,\frac{\partial \mcL(f_{\theta^t}(\S_N),\y_N)}{\partial f_{\theta^t}(\S_N)}\right]\cdot\left[\left\langle\frac{\partial f_{\theta^t}(\cdot)}{\partial \theta^t},\frac{\partial f_{\theta^t}(\S_i)}{\partial \theta^t} \right\rangle\right]_N\nonumber\\
	&=&-\frac{\eta}{NS}\left[\frac{\partial \mcL(f_{\theta^t}(\S_1),\y_1)}{\partial f_{\theta^t}(\S_1)},\dots,\frac{\partial \mcL(f_{\theta^t}(\S_N),\y_N)}{\partial f_{\theta^t}(\S_N)}\right]\cdot\left[K_{\theta^t}(\S_i,\cdot)\right]_N,
\end{eqnarray}
which leads to Equation~\ref{bfparat} expressed as
\begin{eqnarray}
	\frac{\partial f_{\theta^t}}{\partial t}=-\frac{\eta}{NS}\left[\frac{\partial \mcL(f_{\theta^t}(\S_1),\y_1)}{\partial f_{\theta^t}(\S_1)},\dots,\frac{\partial \mcL(f_{\theta^t}(\S_N),\y_N)}{\partial f_{\theta^t}(\S_N)}\right]\cdot\left[K_{\theta^t}(\S_i,\cdot)\right]_N+ o\left(\frac{\partial \theta^t}{\partial t}\right),
\end{eqnarray}
where the symmetric and positive definite $K_{\theta^t}(\S_i,\cdot)\coloneqq\left\langle\frac{\partial f_{\theta^t}(\S_i)}{\partial \theta^t},\frac{\partial f_{\theta^t}(\cdot)}{\partial \theta^t} \right\rangle$ is called the attention neural tangent kernel (ANTK)~\citep{jacot2018neural,yang2019wide,hron2020infinite}. Specifically, ANTK for ${\S}_{(i,:)}$ and ${\S'}_{(j,:)}$ is a scalar $K({\S}_{(i,:)},{\S'}_{(j,:)})=\Big\langle\frac{\partial {f_{\theta^t}(\S)}_{(i,:)}}{\partial \theta^t},\frac{\partial {f_{\theta^t}(\S')}_{(j,:)}}{\partial \theta^t} \Big\rangle$.
Figure~\ref{ntk} illustrates the ANTK computation process, where typically, the length of all training sequences is standardized to the maximum length. In simple terms, examining a model's behavior by focusing on the model itself, rather than its parameters, often involves the use of kernel functions.

The quantity $\frac{\partial f_{\theta^t}(\cdot)}{\partial \theta^t}$, which represents the partial derivative of the ANN with respect to its parameters and appears in $K_{\theta^t}(\S_i,\cdot)=\left\langle\frac{\partial f_{\theta^t}(\S_i)}{\partial \theta^t},\frac{\partial f_{\theta^t}(\cdot)}{\partial \theta^t} \right\rangle$, is determined by both the network architecture and the specific parameters $\theta^t$, but it is independent of the input sequences. In contrast, the term $\frac{\partial f_{\theta^t}(\S_i)}{\partial \theta^t}$ depends not only on the ANN structure and specific $\theta^t$, but also on the input sequence $\S$. When the input for $\frac{\partial f_{\theta^t}(\S_i)}{\partial \theta^t}$ is unspecified, the ANTK simplifies to a general form $K_{\theta^t}(\cdot,\cdot)$. However, when a specific sequence $\S_j$ is provided as the input to $\frac{\partial f_{\theta^t}(\cdot)}{\partial \theta^t}$, the ANTK becomes a matrix defined as $K_{\theta^t}(\S_i,\S_j)=\left\langle\frac{\partial f_{\theta^t}(\S_i)}{\partial \theta^t},\frac{\partial f_{\theta^t}(\S_j)}{\partial \theta^t} \right\rangle$. This formulation aligns with the vector-valued kernel used in functional gradient descent \citep{zhang2023mint}. When the input sequence $\S_i$ is specified, one argument of $K_{\theta^t}$ is fixed, leading the ANN to update along $K_{\theta^t}(\S_i,\cdot)$, with the magnitude of the update determined by $\frac{\partial f_{\theta^t}(\S_i)}{\partial \theta^t}$.
This process reflects the core mechanism of functional gradient descent. In summary, the ANTK and the canonical vector-valued kernel share a consistent mathematical framework and exhibit similar effects on the evolution of the associated ANN. Additionally, Theorem~\ref{ntkconverge} establishes the asymptotic relationship between the ANTK and the canonical kernel used in functional gradient descent.

\subsection{AtteNT Algorithm}\label{app:attent_algo}
\begin{algorithm}[H]
	\caption{AtteNT Algorithm}
        \label{attent_algo}
	{\bfseries Input:} Target mapping $f^*$ realized by a dense set of sequence-property pairs, initial ANN $f_{\theta^0}$, the size of selected training set $m\leq N$, small constant $\epsilon>0$ and maximal iteration number $T$
    	\BlankLine
    	Set $f_{\theta^t}\gets f_{\theta^0}$, $t=0$
    	\BlankLine
    	\While{$t\leq T$ {\rm and} $\left\|\left[f_{\theta^t}(\S_i)-f^*(\S_i)\right]_N\right\|_\mcF\geq\epsilon$}{
    		\BlankLine
    		\textbf{The teacher} selects $m$ teaching sequences:
                \BlankLine
    		\tcc{\small{Sequences associated with the $m$ largest $\|f_{\theta^t}(\S_i) - f^*(\S_i)\|_2$}}
      
                ${\{\S_i\}_m}^*=\underset{\{\S_i\}_m\subseteq\{\S_i\}_N}{\arg\max}\left\|\left[f_{\theta^t}(\S_i)-f^*(\S_i)\right]_m\right\|_\mcF$
                
                \BlankLine
                
    		Provide ${\{\S_i\}_m}^*$ to the attention learner
    		\BlankLine
    		\textbf{The learner} updates $f_{\theta^t}$ based on received ${\{\S_i\}_m}^*$:
                \BlankLine
                \tcp{Parameter-based gradient descent}
    		
    		$\theta^t\gets \theta^t-\frac{\eta}{mS}\sum_{\S_i\in{\{\S_i\}_m}^*}\sum\ind{j}{1}{S}\nabla_{\theta}\mcL({f_{\theta^t}(\S_i)}_{(j,:)},{f^*(\S_i)}_{(j,:)})$
      		\BlankLine
    		Set $t\gets t+1$
    		}
\end{algorithm}

\clearpage
\section{Detailed Proofs}\label{app:dp}

\subsection{Proof of Theorem~\ref{ntkconverge}}\label{pntkconverge} 

By examining the evolution of an ANN through changes in its parameters and from a high-level perspective within the function space, we obtain.
\begin{eqnarray}
	&&-\frac{\eta}{NS}\left[\frac{\partial \mcL(f_{\theta^t}(\S_1),\y_1)}{\partial f_{\theta^t}(\S_1)},\dots,\frac{\partial \mcL(f_{\theta^t}(\S_N),\y_N)}{\partial f_{\theta^t}(\S_N)}\right]\left[K({\S_i},\cdot)\right]_N\nonumber\\
    &=&-\frac{\eta}{NS}\left[\frac{\partial \mcL(f_{\theta^t}(\S_1),\y_1)}{\partial f_{\theta^t}(\S_1)},\dots,\frac{\partial \mcL(f_{\theta^t}(\S_N),\y_N)}{\partial f_{\theta^t}(\S_N)}\right]\cdot\left[\left\langle\frac{\partial f_{\theta^t}(\S_i)}{\partial \theta^t},\frac{\partial f_{\theta^t}(\cdot)}{\partial \theta^t} \right\rangle\right]_N+ o\left(\frac{\partial \theta^t}{\partial t}\right).\nonumber\\
\end{eqnarray}
Upon reorganizing, we derive
\begin{eqnarray}
	-\frac{\eta}{NS}\left[\frac{\partial \mcL(f_{\theta^t}(\S_1),\y_1)}{\partial f_{\theta^t}(\S_1)},\dots,\frac{\partial \mcL(f_{\theta^t}(\S_N),\y_N)}{\partial f_{\theta^t}(\S_N)}\right]\cdot \left[K({\S_i},\cdot)-K_{\theta^t}(\S_i,\cdot)\right]_N=o\left(\frac{\partial \theta^t}{\partial t}\right).
\end{eqnarray}
By integrating the parameter evolution
\begin{eqnarray}
	\frac{\partial\theta^t}{\partial t}=-\eta\frac{\partial\mcL}{\partial\theta^t}=-\frac{\eta}{NS}\left[\frac{\partial \mcL(f_{\theta^t}(\S_1),\y_1)}{\partial f_{\theta^t}(\S_1)},\dots,\frac{\partial \mcL(f_{\theta^t}(\S_N),\y_N)}{\partial f_{\theta^t}(\S_N)}\right]\cdot \left[\frac{\partial f_{\theta^t}(\S_i)}{\partial \theta^t}\right]_N
\end{eqnarray}
into the remainder, we get
\begin{eqnarray}
	&&-\frac{\eta}{NS}\left[\frac{\partial \mcL(f_{\theta^t}(\S_1),\y_1)}{\partial f_{\theta^t}(\S_1)},\dots,\frac{\partial \mcL(f_{\theta^t}(\S_N),\y_N)}{\partial f_{\theta^t}(\S_N)}\right]\cdot \left[K({\S_i},\cdot)-K_{\theta^t}(\S_i,\cdot)\right]_N\nonumber\\
    &=&o\left(-\frac{\eta}{NS}\left[\frac{\partial \mcL(f_{\theta^t}(\S_1),\y_1)}{\partial f_{\theta^t}(\S_1)},\dots,\frac{\partial \mcL(f_{\theta^t}(\S_N),\y_N)}{\partial f_{\theta^t}(\S_N)}\right]\cdot \left[\frac{\partial f_{\theta^t}(\S_i)}{\partial \theta^t}\right]_N\right).
\end{eqnarray}
When training an ANN with a convex loss $\mcL$, which is convex in terms of $f_\theta$ but not necessarily in terms of $\theta$, the following limit holds for the vector:  $\lim_{t\to\infty}\left[\frac{\partial \mcL(f_{\theta^t}(\S_1),\y_1)}{\partial f_{\theta^t}(\S_1)},\dots,\frac{\partial \mcL(f_{\theta^t}(\S_N),\y_N)}{\partial f_{\theta^t}(\S_N)}\right]=\bm{0}$. Since the right-hand side of this equation is a higher-order infinitesimal relative to the left, maintaining this equality results in the conclusion that
\begin{eqnarray}
	\lim_{t\to\infty} \ \left[K({\S_i},\cdot)-K_{\theta^t}(\S_i,\cdot)\right]_N=\bm{0}~.
\end{eqnarray}
This suggests that for each training point, \ie, input sequence $\S\in\{\S_i\}_N$, ANTK converges pointwise to the canonical kernel.

$\blacksquare$

\subsection{Proof of Proposition~\ref{slr}} \label{pslr}

Referring to the definition of the Fréchet derivative in Definition~\ref{fd}, the convexity of $\mcL$ implies that
\begin{eqnarray}
	\frac{\partial\mcL}{\partial t}\leq\underbrace{\left\langle\frac{\partial\mcL}{\partial f_{\theta^{t+1}}},\frac{f_{\theta^t}}{\partial t}\right\rangle_\mathcal{H}}_{\Upsilon}.
\end{eqnarray}
By computing the Fréchet derivative of $\frac{\partial\mcL}{\partial f_{\theta^{t+1}}}$ and the evolution of $f_{\theta^t}$, the term on the right-hand side, $\Upsilon$, can be expressed as
\begin{eqnarray}\label{eqxi}
	\Upsilon&=&\left\langle\mcG^{t+1},-\eta\mcG^t\right\rangle_{\mathcal{H}}\nonumber\\
	&=&-\frac{\eta}{N^2S^2}\left\langle\left[\frac{\partial \mcL(f_{\theta^{t+1}}(\S_1),\y_1)}{\partial f_{\theta^{t+1}}(\S_1)},\dots,\frac{\partial \mcL(f_{\theta^{t+1}}(\S_N),\y_N)}{\partial f_{\theta^{t+1}}(\S_N)}\right]\cdot \left[K_{\S_i}\right]_N,\right.\nonumber\\
    &&\qquad\qquad\left.[K_{\S_i}]\trsp_N\cdot\left[\frac{\partial \mcL(f_{\theta^t}(\S_1),\y_1)}{\partial f_{\theta^t}(\S_1)},\dots,\frac{\partial \mcL(f_{\theta^t}(\S_N),\y_N)}{\partial f_{\theta^t}(\S_N)}\right]\trsp\right\rangle_{\mathcal{H}}\nonumber\\
	&=&-\frac{\eta}{N^2S^2}\left[\frac{\partial \mcL(f_{\theta^{t+1}}(\S_1),\y_1)}{\partial f_{\theta^{t+1}}(\S_1)},\dots,\frac{\partial \mcL(f_{\theta^{t+1}}(\S_N),\y_N)}{\partial f_{\theta^{t+1}}(\S_N)}\right]\cdot \left\langle\left[K_{\S_i}\right]_N,[K_{\S_i}]\trsp_N\right\rangle_{\mathcal{H}}\nonumber\\
    &&\qquad\qquad\cdot\left[\frac{\partial \mcL(f_{\theta^t}(\S_1),\y_1)}{\partial f_{\theta^t}(\S_1)},\dots,\frac{\partial \mcL(f_{\theta^t}(\S_N),\y_N)}{\partial f_{\theta^t}(\S_N)}\right]\trsp\nonumber\\
	&=&-\frac{\eta}{NS}\left[\frac{\partial \mcL(f_{\theta^{t+1}}(\S_1),\y_1)}{\partial f_{\theta^{t+1}}(\S_1)},\dots,\frac{\partial \mcL(f_{\theta^{t+1}}(\S_N),\y_N)}{\partial f_{\theta^{t+1}}(\S_N)}\right]\bar{\bm{K}}\left[\frac{\partial \mcL(f_{\theta^t}(\S_1),\y_1)}{\partial f_{\theta^t}(\S_1)},\dots,\frac{\partial \mcL(f_{\theta^t}(\S_N),\y_N)}{\partial f_{\theta^t}(\S_N)}\right]\trsp,\nonumber\\
\end{eqnarray}
where $\bar{\bm{K}}=\bm{K}/(NS)$, and $\bm{K}$ is an $NS\times NS$ symmetric, positive definite block matrix with elements $K(\S_i,\S_j)$ positioned in the $i$-th row and $j$-th column block. For convenience, we use a simplified column vector notation $\left[\partial_{f_{\theta^\square}}\mcL(f_{\theta^\square};\S_i)\right]_N\coloneqq\left[\partial_{f_{\theta^\square}}\mcL(f_{\theta^\square};\S_1),\dots,\partial_{f_{\theta^\square}}\mcL(f_{\theta^\square};\S_N)\right]\trsp$ with $\partial_{f_{\theta^\square}}\mcL(f_{\theta^\square};\S_i)\coloneqq\frac{\partial \mcL(f_{\theta^\square}(\S_i),\y_i)}{\partial f_{\theta^\square}(\S_i)}$. The last term in Equation~\ref{eqxi} can then be rewritten as
\begin{eqnarray}\label{epxi}
	&&-\frac{\eta}{NS}\left[\partial_{f_{\theta^t}}\mcL(f_{\theta^t};\S_i)\right]\trsp_N\bar{\bm{K}}\left[\partial_{f_{\theta^{t+1}}}\mcL(f_{\theta^{t+1}};\S_i)\right]_N\nonumber\\
	&=&-\frac{\eta}{NS}\left[\partial_{f_{\theta^t}}\mcL(f_{\theta^t};\S_i)\right]\trsp_N\bar{\bm{K}}\left(\left[\partial_{f_{\theta^{t+1}}}\mcL(f_{\theta^{t+1}};\S_i)\right]_N+\left[\partial_{f_{\theta^t}}\mcL(f_{\theta^t};\S_i)\right]_N-\left[\partial_{f_{\theta^t}}\mcL(f_{\theta^t};\S_i)\right]_N\right)\nonumber\\
	&=&-\frac{\eta}{NS}\left[\partial_{f_{\theta^t}}\mcL(f_{\theta^t};\S_i)\right]\trsp_N\bar{\bm{K}}\left[\partial_{f_{\theta^t}}\mcL(f_{\theta^t};\S_i)\right]_N\nonumber\\
    &&-\frac{\eta}{NS}\left[\partial_{f_{\theta^t}}\mcL(f_{\theta^t};\S_i)\right]\trsp_N\bar{\bm{K}}\left(\left[\partial_{f_{\theta^{t+1}}}\mcL(f_{\theta^{t+1}};\S_i)\right]_N-\left[\partial_{f_{\theta^t}}\mcL(f_{\theta^t};\S_i)\right]_N\right)\nonumber\\
	&=&-\frac{\eta}{NS}\left[\partial_{f_{\theta^t}}\mcL(f_{\theta^t};\S_i)\right]\trsp_N\bar{\bm{K}}\left[\partial_{f_{\theta^t}}\mcL(f_{\theta^t};\S_i)\right]_N\nonumber\\
	&&+\frac{\eta}{NS}\left(\left[\partial_{f_{\theta^{t+1}}}\mcL(f_{\theta^{t+1}};\S_i)\right]\trsp_N-\left[\partial_{f_{\theta^t}}\mcL(f_{\theta^t};\S_i)\right]\trsp_N-\left[\partial_{f_{\theta^{t+1}}}\mcL(f_{\theta^{t+1}};\S_i)\right]\trsp_N\right)\nonumber\\
    &&\cdot\bar{\bm{K}}\cdot\left(\left[\partial_{f_{\theta^{t+1}}}\mcL(f_{\theta^{t+1}};\S_i)\right]_N-\left[\partial_{f_{\theta^t}}\mcL(f_{\theta^t};\S_i)\right]_N\right)~.
\end{eqnarray}
The last term in Equation~\ref{epxi} above can be expanded to
\begin{eqnarray}\label{lstepxi}
	&&\frac{\eta}{NS}\left(\left[\partial_{f_{\theta^{t+1}}}\mcL(f_{\theta^{t+1}};\S_i)\right]\trsp_N-\left[\partial_{f_{\theta^t}}\mcL(f_{\theta^t};\S_i)\right]\trsp_N-\left[\partial_{f_{\theta^{t+1}}}\mcL(f_{\theta^{t+1}};\S_i)\right]\trsp_N\right)\nonumber\\
    &&\cdot\bar{\bm{K}}\left(\left[\partial_{f_{\theta^{t+1}}}\mcL(f_{\theta^{t+1}};\S_i)\right]_N-\left[\partial_{f_{\theta^t}}\mcL(f_{\theta^t};\S_i)\right]_N\right)\nonumber\\
	&=&\frac{\eta}{NS}\left(\left[\partial_{f_{\theta^{t+1}}}\mcL(f_{\theta^{t+1}};\S_i)\right]_N-\left[\partial_{f_{\theta^t}}\mcL(f_{\theta^t};\S_i)\right]_N\right)\trsp\bar{\bm{K}}\left(\left[\partial_{f_{\theta^{t+1}}}\mcL(f_{\theta^{t+1}};\S_i)\right]_N-\left[\partial_{f_{\theta^t}}\mcL(f_{\theta^t};\S_i)\right]_N\right)\nonumber\\
	&&-\frac{\eta}{NS}\left[\partial_{f_{\theta^{t+1}}}\mcL(f_{\theta^{t+1}};\S_i)\right]\trsp_N\bar{\bm{K}}\left(\left[\partial_{f_{\theta^{t+1}}}\mcL(f_{\theta^{t+1}};\S_i)\right]_N-\left[\partial_{f_{\theta^t}}\mcL(f_{\theta^t};\S_i)\right]_N\right)\nonumber\\
	&=&\frac{\eta}{NS}\left[\partial_{f_{\theta^{t+1}}}\mcL(f_{\theta^{t+1}};\S_i)-\partial_{f_{\theta^t}}\mcL(f_{\theta^t};\S_i)\right]_N\trsp\bar{\bm{K}}\left[\partial_{f_{\theta^{t+1}}}\mcL(f_{\theta^{t+1}};\S_i)-\partial_{f_{\theta^t}}\mcL(f_{\theta^t};\S_i)\right]_N\nonumber\\
	&&-\frac{\eta}{NS}\left(\left[\partial_{f_{\theta^{t+1}}}\mcL(f_{\theta^{t+1}};\S_i)\right]_N-\frac{1}{2}\left[\partial_{f_{\theta^t}}\mcL(f_{\theta^t};\S_i)\right]_N\right)\trsp\bar{\bm{K}}\left(\left[\partial_{f_{\theta^{t+1}}}\mcL(f_{\theta^{t+1}};\S_i)\right]_N-\frac{1}{2}\left[\partial_{f_{\theta^t}}\mcL(f_{\theta^t};\S_i)\right]_N\right)\nonumber\\
	&&+\frac{\eta}{4NS}\left[\partial_{f_{\theta^t}}\mcL(f_{\theta^t};\S_i)\right]\trsp_N\bar{\bm{K}}\left[\partial_{f_{\theta^t}}\mcL(f_{\theta^t};\S_i)\right]_N~.
\end{eqnarray}
Given that $\bar{\bm{K}}$ is positive definite, it follows that $$\frac{\eta}{NS}\left(\left[\partial_{f_{\theta^{t+1}}}\mcL(f_{\theta^{t+1}};\S_i)\right]_N-\frac{1}{2}\left[\partial_{f_{\theta^t}}\mcL(f_{\theta^t};\S_i)\right]_N\right)\trsp\bar{\bm{K}}\left(\left[\partial_{f_{\theta^{t+1}}}\mcL(f_{\theta^{t+1}};\S_i)\right]_N-\frac{1}{2}\left[\partial_{f_{\theta^t}}\mcL(f_{\theta^t};\S_i)\right]_N\right)$$ is a non-negative term. Therefore, by merging Equations \ref{eqxi}, \ref{epxi}, and \ref{lstepxi}, we derive
\begin{eqnarray}\label{xileq}
	\Upsilon&\leq&-\frac{3\eta}{4NS}\underbrace{\left[\partial_{f_{\theta^t}}\mcL(f_{\theta^t};\S_i)\right]\trsp_N\bar{\bm{K}}\left[\partial_{f_{\theta^t}}\mcL(f_{\theta^t};\S_i)\right]_N}_{\Phi}\nonumber\\
	&&+\frac{\eta}{NS}\underbrace{\left[\partial_{f_{\theta^{t+1}}}\mcL(f_{\theta^{t+1}};\S_i)-\partial_{f_{\theta^t}}\mcL(f_{\theta^t};\S_i)\right]_N\trsp\bar{\bm{K}}\left[\partial_{f_{\theta^{t+1}}}\mcL(f_{\theta^{t+1}};\S_i)-\partial_{f_{\theta^t}}\mcL(f_{\theta^t};\S_i)\right]_N}_{\Psi}~.\nonumber\\
\end{eqnarray}

Based on the definition of the evaluation functional and the assumption that $\mcL$ is Lipschitz smooth with a constant $\tau>0$, the term $\Psi$ in the final part of Equation~\ref{xileq} is bounded above as follows:
\begin{eqnarray}\label{ls}
	\Psi&=&\left[\partial_{f_{\theta^{t+1}}}\mcL(f_{\theta^{t+1}};\S_i)-\partial_{f_{\theta^t}}\mcL(f_{\theta^t};\S_i)\right]_N\trsp\bar{\bm{K}}\left[\partial_{f_{\theta^{t+1}}}\mcL(f_{\theta^{t+1}};\S_i)-\partial_{f_{\theta^t}}\mcL(f_{\theta^t};\S_i)\right]_N\nonumber\\
	&=&\left[E_{\S_i}\left(\frac{\partial\mcL(f_{\theta^{t+1}})}{\partial f_{\theta^{t+1}}}-\frac{\partial\mcL(f_{\theta^t})}{\partial f_{\theta^t}}\right)\right]\trsp_N\bar{\bm{K}}\left[E_{\S_i}\left(\frac{\partial\mcL(f_{\theta^{t+1}})}{\partial f_{\theta^{t+1}}}-\frac{\partial\mcL(f_{\theta^t})}{\partial f_{\theta^t}}\right)\right]_N\nonumber\\
	&\leq&\tau^2\left[E_{\S_i}\left(f_{\theta^{t+1}}-f_{\theta^t}\right)\right]\trsp_N\bar{\bm{K}}\left[E_{\S_i}\left(f_{\theta^{t+1}}-f_{\theta^t}\right)\right]_N\nonumber\\
	&=&\tau^2\left\langle\left(f_{\theta^{t+1}}-f_{\theta^t}\right),\left[K_{\S_i}\right]\trsp_N\right\rangle_{\mathcal{H}}\cdot\bar{\bm{K}}\cdot\left\langle\left[K_{\S_i}\right]_N,\left(f_{\theta^{t+1}}-f_{\theta^t}\right)\right\rangle_{\mathcal{H}}\nonumber\\
	&=&{\eta}^2\tau^2\cdot\left[\partial_{f_{\theta^t}}\mcL(f_{\theta^t};\S_i)\right]_N\trsp\frac{\left\langle\left[K_{\S_i}\right]_N,[K_{\S_i}]\trsp_N\right\rangle_{\mathcal{H}}}{NS}\cdot\bar{\bm{K}}\cdot\frac{\left\langle\left[K_{\S_i}\right]_N,[K_{\S_i}]\trsp_N\right\rangle_{\mathcal{H}}}{NS}\cdot\left[\partial_{f_{\theta^t}}\mcL(f_{\theta^t};\S_i)\right]_N~.\nonumber\\
\end{eqnarray}
Given that the canonical kernel is bounded above by a constant $\gamma>0$, we have 
\begin{eqnarray}
	\left\langle\left[K_{\S_i}\right]_N,[K_{\S_i}]\trsp_N\right\rangle_{\mathcal{H}}\leq\gamma\left\langle \1_{NS},\1_{NS}\trsp\right\rangle,\nonumber
\end{eqnarray}
and 
\begin{eqnarray}
	\bar{\bm{K}}\leq\frac{\gamma}{NS}\left\langle \1_{NS},\1_{NS}\trsp\right\rangle~. \nonumber
\end{eqnarray} 
Therefore, $\Phi$ is bounded above by
\begin{eqnarray}\label{bk1}
	\Phi&\leq&\frac{\gamma}{NS}\left\langle\left[\partial_{f_{\theta^t}}\mcL(f_{\theta^t};\S_i)\right]\trsp_N,\1_{NS}\right\rangle\left\langle\1_{NS}\trsp,\left[\partial_{f_{\theta^t}}\mcL(f_{\theta^t};\S_i)\right]_N\right\rangle\nonumber\\
	&=&\frac{\gamma}{NS}\left(\sum_{i=1}^N\sum_{j=1}^S\partial_{f_{\theta^t}}\mcL(f_{\theta^t};{\S_i}_{(j,:)})\right)^2.
\end{eqnarray}
Moreover, the last term in Equation~\ref{ls} is also bounded above:
\begin{eqnarray}\label{bk2}
	&&{\eta}^2\tau^2\cdot\left[\partial_{f_{\theta^t}}\mcL(f_{\theta^t};\S_i)\right]_N\trsp\frac{\left\langle\left[K_{\S_i}\right]_N,[K_{\S_i}]\trsp_N\right\rangle_{\mathcal{H}}}{NS}\cdot\bar{\bm{K}}\cdot\frac{\left\langle\left[K_{\S_i}\right]_N,[K_{\S_i}]\trsp_N\right\rangle_{\mathcal{H}}}{NS}\cdot\left[\partial_{f_{\theta^t}}\mcL(f_{\theta^t};\S_i)\right]_N\nonumber\\
	&\leq&{\eta}^2\tau^2\left[\frac{\gamma}{NS}\sum_{i=1}^N\sum_{j=1}^S\partial_{f_{\theta^t}}\mcL(f_{\theta^t};{\S_i}_{(j,:)})\right]\trsp\cdot\bar{\bm{K}}\cdot\left[\frac{\gamma}{NS}\sum_{i=1}^N\sum_{j=1}^S\partial_{f_{\theta^t}}\mcL(f_{\theta^t};{\S_i}_{(j,:)})\right]_N\nonumber\\
	&\leq&\frac{{\eta}^2\tau^2\gamma^3}{NS}\left\langle\left[\frac{1}{NS}\sum_{i=1}^N\sum_{j=1}^S\partial_{f_{\theta^t}}\mcL(f_{\theta^t};{\S_i}_{(j,:)})\right]\trsp_N,\1_{NS}\right\rangle\left\langle\1_{NS}\trsp,\left[\frac{1}{NS}\sum_{i=1}^N\sum_{j=1}^S\partial_{f_{\theta^t}}\mcL(f_{\theta^t};{\S_i}_{(j,:)})\right]_N\right\rangle\nonumber\\
	&=&\frac{{\eta}^2\tau^2\gamma^3}{NS}\left(\sum_{i=1}^N\sum_{j=1}^S\partial_{f_{\theta^t}}\mcL(f_{\theta^t};{\S_i}_{(j,:)})\right)^2.
\end{eqnarray}
Thus, by combining Equations~\ref{xileq}, \ref{ls}, \ref{bk1}, and \ref{bk2}, we get
\begin{eqnarray}
	\Upsilon\leq-\eta\gamma\left(\frac{3}{4}-{\eta}^2\tau^2\gamma^2\right)\left(\frac{1}{NS}\sum_{i=1}^N\sum_{j=1}^S\partial_{f_{\theta^t}}\mcL(f_{\theta^t};{\S_i}_{(j,:)})\right)^2,
\end{eqnarray}
which means
\begin{eqnarray}
	\frac{\partial \mcL}{\partial t}\leq\Upsilon\leq-\eta\gamma\left(\frac{3}{4}-{\eta}^2\tau^2\gamma^2\right)\left(\frac{1}{NS}\sum_{i=1}^N\sum_{j=1}^S\partial_{f_{\theta^t}}\mcL(f_{\theta^t};{\S_i}_{(j,:)})\right)^2.
\end{eqnarray}
Hence, if $\eta\leq\frac{1}{2\tau\gamma}$, it follows that
\begin{eqnarray}
	\frac{\partial \mcL}{\partial t}\leq -\frac{\eta\gamma}{2}\left(\frac{1}{NS}\sum_{i=1}^N\sum_{j=1}^S\partial_{f_{\theta^t}}\mcL(f_{\theta^t};{\S_i}_{(j,:)})\right)^2=-\frac{\eta\gamma}{2}\left(\frac{1}{NS}\sum_{i=1}^N\sum_{j=1}^S\frac{\partial \mcL\left({f_{\theta^t}(\S_i)}_{(j,:)},{\y_i}_{(j,:)}\right)}{\partial {f_{\theta^t}(\S_i)}_{(j,:)}}\right)^2.
\end{eqnarray}

$\blacksquare$

\clearpage
\section{Experiment Details}
\label{app:ade}

\subsection{LLMs Training Setting}

All experiments were conducted on 4 NVIDIA A100 (80GB) GPUs. We employ LoRA fine-tuning following the Alpaca~\citep{alpaca} and Pizza~\citep{meng2024pissa} implementation strategy. Specifically, we optimize with AdamW using a batch size of 128, a learning rate of $2\text{e}{-5}$, cosine annealing scheduling~\citep{DBLP:journals/corr/LoshchilovH16a}, and a warmup ratio of 0.03, without weight decay. The training objective computes loss only over responses from the selected datasets. We configure LoRA with lora\_alpha=lora\_r, set lora\_dropout, and insert adapters into all linear layers of the base model. Both the backbone and adapters are trained in Float32 precision.

\paragraph{AtteNT Setting} In this experiment, we adopt a straightforward variant of AtteNT: the model is trained on the full dataset during the first epoch, after which only the AtteNT selected subset is used in subsequent epochs. Selection is guided by the per-sample loss scores within each epoch, effectively directing the model's attention toward harder examples. Since pretrained models already perform well on most instances, emphasizing more challenging data in later epochs is expected to yield greater fine-tuning benefits. Following prior findings in Rho-1~\citep{lin2024not}, we set the selection ratio to 70\%.

\paragraph{Dataset Building} The ImageNetS50 dataset is derived from the ImageNet-1k benchmark, and its construction requires access to a local copy of ImageNet-1k. Following the official repository~\citep{imagenetS}, we generate ImageNetS50 by running \texttt{data\_preparation.sh} with the option \texttt{--mode=50}. Semantic segmentation annotations are obtained using \texttt{datapreparation\_anno.sh}. For depth annotations, we employ the Mask2Former~\citep{mask2former} framework, utilizing its released code and pretrained models to generate pseudo-labels. In addition, we directly download the full NYUv2 dataset, where the official semantic segmentation and depth test sets are used for evaluation.

\subsection{ViTs Training Setting} 
We adopt ViT-B~\citep{dosovitskiyimage} with a $16\times16$ patch size as the backbone for our MAE experiments and evaluate performance on ImageNet-S50. Training is performed using AdamW with a base learning rate of $1\text{e}{-4}$ and weight decay of 0.05. The learning rate is linearly warmed up for 40 epochs, followed by cosine decay scheduling~\citep{DBLP:journals/corr/LoshchilovH16a}. We train with a batch size of 2048 on 4 A100 GPUs, leveraging automatic mixed precision for efficiency. Data augmentation is limited to standard transformations: random cropping with scale sampled from [0.2, 1.0] and aspect ratio from [0.75, 1.33], resizing to $224\times224$, and random horizontal flipping with probability 0.5. A full specification of hyperparameters for pretraining and fine-tuning is provided in Tables~\ref{multimae-pretrain} and~\ref{multimae-finetune}.

\begin{table}[h]
\centering
\caption{Hyperparameters for pre-training Multi-Modal MAE.}
\label{multimae-pretrain}
\resizebox{.8\linewidth}{!}{%
\begin{tabular}{l@{\hspace{1.2cm}}c@{\hspace{1.2cm}}c}
\toprule
{\bf Hyperparam} &
{\bf Baseline} & {\bf AtteNT}\\ 
\hline 
Batch Size & 2048 & 2048  \\
Learning Rate & 1e-4 & 1e-4 \\
Min Learning Rate & 1e-6 & 1e-6 \\
Weight Decay & 0.05 & 0.05  \\
Adamw $\epsilon$ & 1e-8 & 1e-8 \\
Adamw $\beta_1$ & 0.9  & 0.9 \\
Adamw $\beta_2$ & 0.95 & 0.95 \\
Epoch & 800 & 800 \\
Warm up Epoch & 40 &40 \\
Learning Rate Schedule & cosine decay & cosine decay\\
Non-masked tokens & 98 & 98 \\
Input resolution & 224$\times$224 & 224$\times$224 \\
Augmentation & RandomResizeCrop & RandomResizeCrop \\
Dropout & 0.0 &0.0 \\
Patch Size & 16 & 16 \\
Selection & \{None\} & \{Random, Hard, Soft\}\\
\bottomrule
\end{tabular}
}
\end{table}


\begin{table}[h]
\centering
\caption{Hyperparameters for fine-tuning Multi-Modal MAE on various downtasks. The augmentation strategy LSJ is large scale jittering~\citep{LSJ}. We use drop path~\citep{drop_path} in classification and semantic segmentation tasks.}
\label{multimae-finetune}
\resizebox{.8\linewidth}{!}{%
\begin{tabular}{l@{\hspace{1.5cm}}c@{\hspace{1cm}}c@{\hspace{1cm}}c}
\toprule
{\bf Hyperparam} & 
{\bf ImageNetS50} & 
{\bf NYUv2(S)} & {\bf NYUv2(D)} \\ 
\hline 
Epoch & 100 & 100 & 2000 \\
Warm up Epoch & 5  & 20 & 100 \\
Batch Size & 1024 &  1024 &2048 \\
Learning Rate & 4e-3  & 1e-4 & 1e-4 \\
Min Learning Rate & 1e-6  & 1e-6 & 0\\
Weight Decay & 0.05 & 0.05 & 1e-4\\
Adamw $\beta_1$ & 0.9  & 0.9  & 0.9\\
Adamw $\beta_2$ & 0.999 & 0.999  & 0.999\\
Layer Decay & 0.65 & 0.75 & 0.75 \\
Patch Size & 16  & 16 & 16 \\
Drop path &0.1  &0.1 & /\\
LR Schedule & cosine decay &cosine decay & cosine decay\\
Input resolution & 224$\times$224 & 224$\times$224  & 256$\times$256 \\
Augmentation & Rand(9, 0.5) & LSJ & LSJ \\
\bottomrule
\end{tabular}
}
\end{table}

\paragraph{AtteNT Setting}

We employ an enhanced AtteNT strategy that dynamically selects training data based on per-sample loss scores. Specifically, data selection in the first epoch is guided by each sample's initial loss, and the selection is periodically updated by recomputing loss scores after fixed intervals. This updated subset is then used to initiate the next training stage. Moreover, we incorporate a dynamic selection ratio from 20\% to 80\%, following the adaptive scheme proposed by~\citep{zhang2023nonparametric}. As demonstrated in Section \ref{sec:Ablation}, this approach achieves a favorable trade-off between efficiency and performance.

\paragraph{Dataset Building} 

All datasets used in our experiments are available on HuggingFace:
\begin{itemize}
    \item Mathematical Reasoning: Training on \texttt{meta-math/MetaMathQA}; evaluation on \texttt{openai/gsm8k} and \texttt{hendrycks/MATH}.
    \item Code Generation: Training on \texttt{m-a-p/CodeFeedback} (restricted to Python samples); evaluation on \texttt{openai/humaneval} and \texttt{google/mbpp}.
    \item Multi-Turn Dialogue: Training on \texttt{WizardLM/evol\_instruct\_196k}; evaluation on \texttt{lmsys/mt-bench}.
\end{itemize}

\clearpage

\section{Additional Experiments}\label{supp:addition_experiments}

\subsection{Ablation of Sample Ratio}\label{supp:sample_ratio}

\begin{table}[ht]
\centering
\caption{Sample Ratio Study of ViT models.}
\label{tab:sample_ratio}
\begin{tabular}{l|ccccccc}
\toprule
\diagbox{\textbf{Tasks}}{\textbf{Ratio}} & \textbf{100} & \textbf{90} & \textbf{80} & \textbf{70} & \textbf{60} & \textbf{50} & \textbf{40} \\
\hline
ImageNetS50    & 92.2 & 91.8 & 91.4 & 85.6 & 78.8 & 64.5 & 58.2 \\
\hline
NYUv2(S) & 51.9 & 51.1 & 50.2 & 46.8 & 38.2 & 29.1 & 21.9 \\
\hline
NYUv2(D)  & 52.1 & 52.2 & 51.6 & 48.3 & 42.4 & 36.3 & 30.5 \\
\bottomrule
\end{tabular}
\end{table}

\begin{figure}[ht]
\centering
\includegraphics[width=0.8\textwidth]{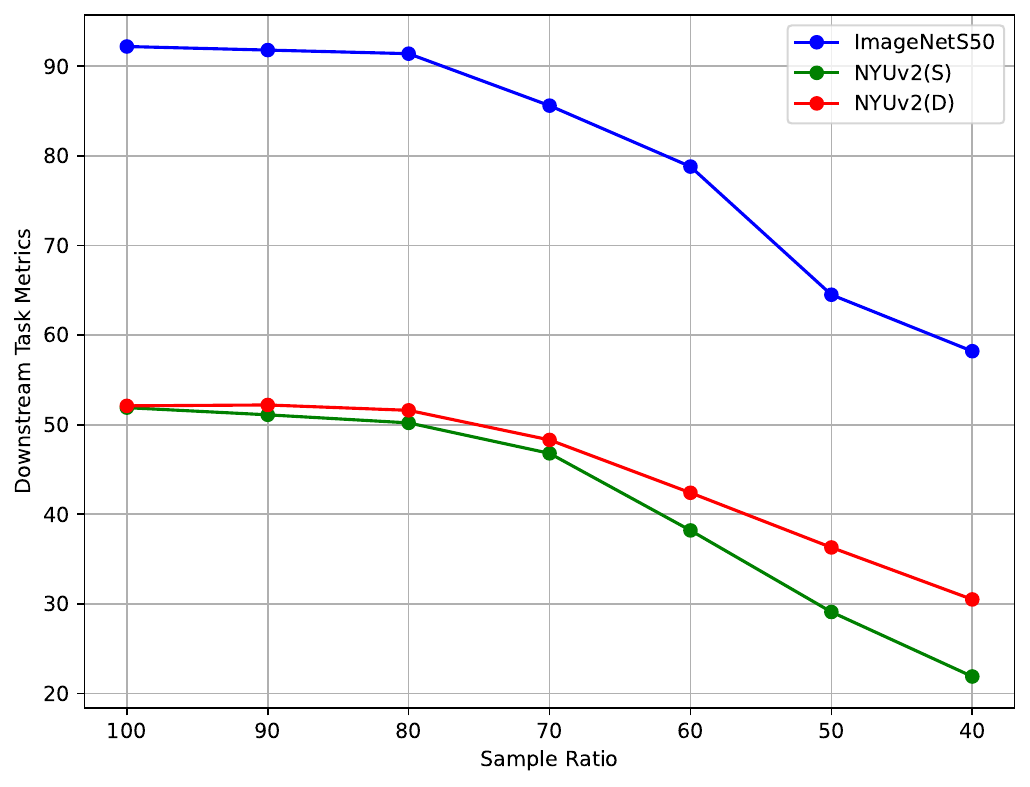}
\caption{Downstream Task Performance vs Sample Ratio.}
\label{fig:sample_ratio}
\end{figure}

In this section, we investigate the impact of the AtteNT algorithm's sample ratio on downstream tasks. In Table~\ref{tab:sample_ratio}, we compare the results based on the ViT model using different fixed sample ratios.

As we can see from Fig~\ref{fig:sample_ratio}, there is a noticeable drop in performance around the 80\% selection ratio. This suggests that, during training, a portion of the data remains relatively unchanged and contributes less to performance when its selection ratio falls below a certain threshold~\citep{lin2024not, katharopoulos2018not}. When the ratio of unselected samples during training is lower than this threshold, the model can maintain its performance. Interestingly, a small data drop can even act as a form of noise reduction.

\subsection{Comparison to Established Methods}\label{supp:comparison}

\begin{table}[ht]
\centering
\caption{Performance Comparison of Different Methods.}
\label{tab:Comparison_with_others}
\resizebox{\linewidth}{!}{%
\begin{tabular}{lcccc}
\toprule
\textbf{Methods} & \textbf{Time(↓)} & \textbf{ImageNetS50(↑)} & \textbf{NYUv2(S)(↑)} & \textbf{NYUv2(D)(↑)} \\
\hline
AtteNT(Ours) & \textbf{980m} & \textbf{92.3} & \textbf{52.6} & \textbf{57.2} \\
\hline
Class Weight Sampling & 1108m & 90.4 & 48.2 & 52.0 \\
\hline
Fixed Weight Sampling & 1065m & 89.6 & 49.7 & 54.6 \\
\hline
GradNorm Sampling~\citep{chen2018gradnorm} & 1112m & 91.9 & 52.4 & 55.8 \\
\bottomrule
\end{tabular}
}
\end{table}

As shown in Table~\ref{tab:Comparison_with_others}, we also performed experiments comparing AtteNT with three simple yet representative sample-selection baselines to compare with traditional greedy algorithm: The first method, Class-Weight Sampling, assigns sampling weights inversely proportional to the number of samples per class, aiming to encourage the model to treat all classes more equally (sampling weight = 1 / class frequency). The second, Fixed-Weight Sampling, assigns fixed sampling ratios based on prior beliefs about task difficulty. In our setting, we consider the classification task easier than semantic segmentation and depth estimation, so we reduce the sampling rate for the RGB modality and set fixed sampling weights to 1 : 2 : 2 (RGB : SemSeg : Depth). The third method, GradNorm Sampling~\citep{chen2018gradnorm}, dynamically adjusts the sampling weights for data groups (RGB, SemSeg, Depth) based on their gradient contributions during training. All methods were trained for 800 epochs, with the total sampling budget fixed at 70\% for each baseline.

Across all comparisons, AtteNT consistently achieves higher efficiency and stronger predictive performance. These results indicate that AtteNT’s gains do not arise from generic greedy sampling heuristics, but from its principled nonparametric teaching mechanism, which adapts to model uncertainty and task interactions more effectively than existing selection strategies.

\subsection{Visualizing NTK Analysis}\label{supp:NTK}

\begin{figure}[ht]
\centering
\includegraphics[width=\textwidth]{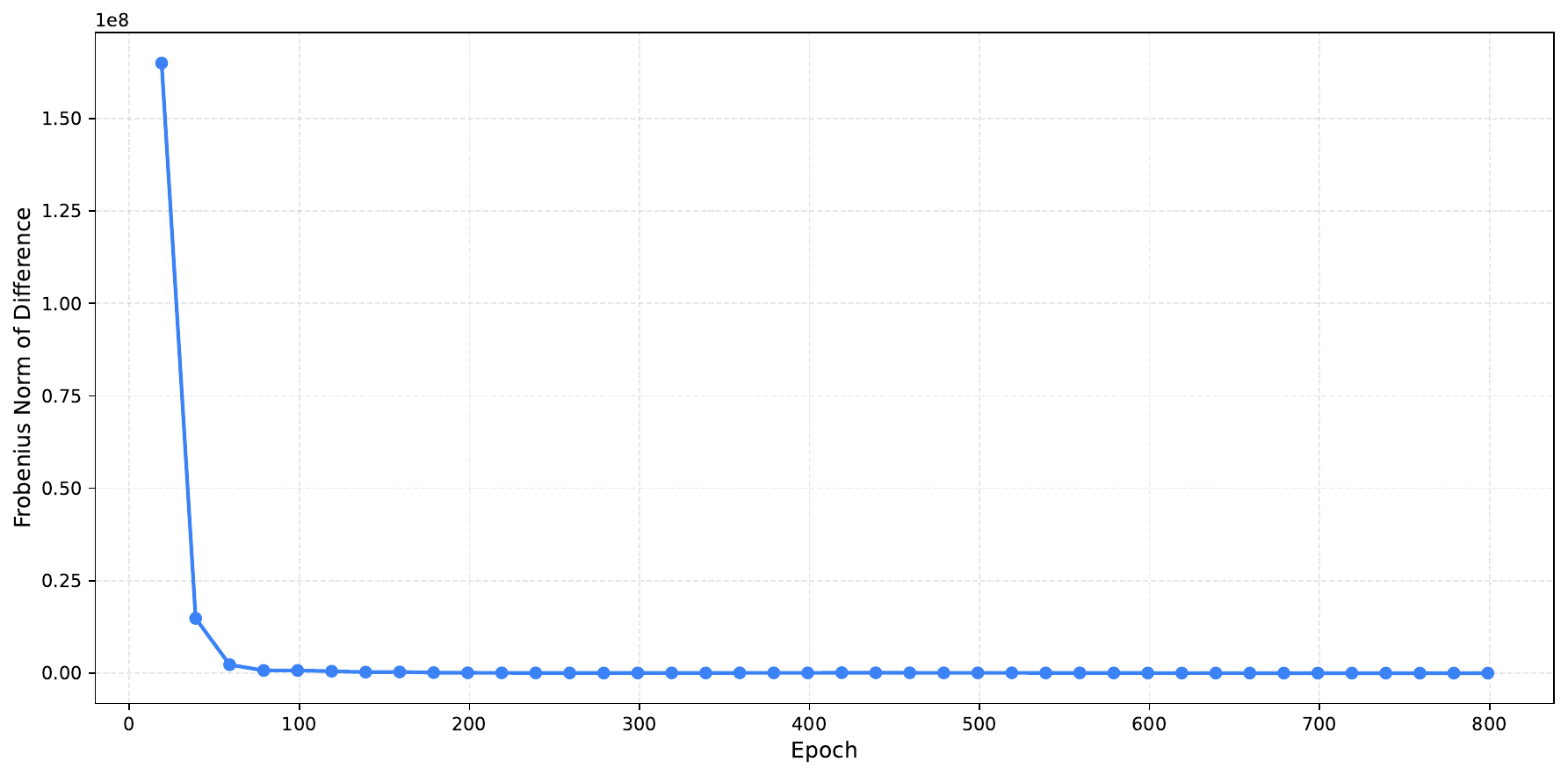}
\caption{Frobenius norm of the difference between the empirical NTK at different training steps and the canonical kernel.}
\label{fig:ntk_difference_plot}
\end{figure}

\begin{figure}[ht]
\centering
\includegraphics[width=\textwidth]{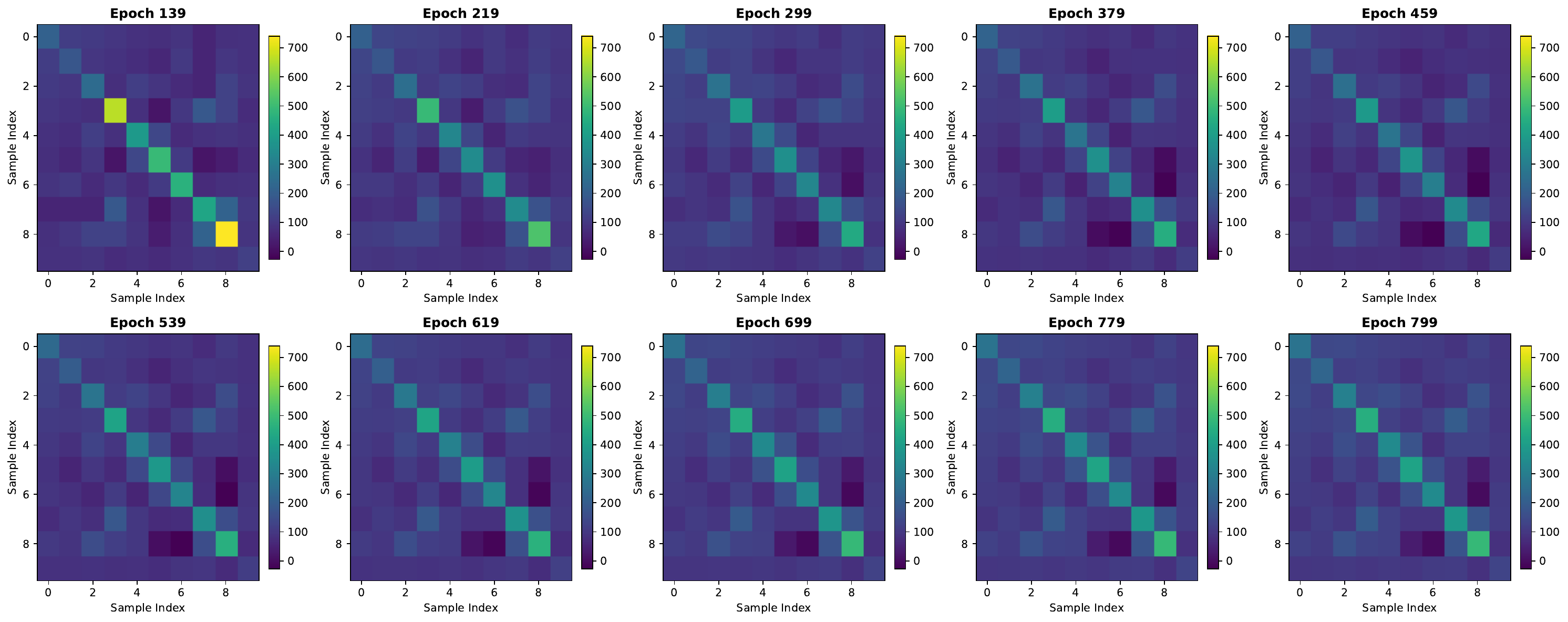}
\caption{Evolution of the empirical $10\times 10$ NTK matrix during training. Color represents value $K_{\theta^t}(\S_i, \S_j)$. The matrix stabilizes visually after ~200 epochs and shows negligible changes thereafter.}
\label{fig:ntk_heatmaps}
\end{figure}

To empirically confirm that the neural tangent kernel quickly stabilizes in real vision transformer training, we track the NTK on 10 fixed training points during an 800-epoch run of the Multi-Modal MAE backbone:

\begin{itemize}
    \item Figure~\ref{fig:ntk_difference_plot}: Frobenius norm of the difference between the empirical NTK at epoch and the canonical kernel. The difference falls sharply within the first ~50 epochs and stays near zero thereafter.
    \item Figure~\ref{fig:ntk_heatmaps}: Heatmaps of the $10\times 10$ NTK at selected checkpoints. A clear pattern is already visible at epoch 139, and from epoch 219 onward the heatmaps are virtually identical and remain unchanged through epoch 799. 
\end{itemize}

These quantitative and qualitative results jointly show that the empirical NTK converges extremely rapidly (within ~50–200 epochs) and remains close to the canonical kernel for the rest of training.

\section{The Use of Large Language Models (LLMs)}
We used large language models for language polishing, such as grammar and phrasing. And we also use AI to assist with code completion. All research ideas, methods, analyses, figures, tables, and conclusions were solely developed by the authors. The authors take full responsibility for all content.

\end{document}